\documentclass[journal]{IEEEtran}
\usepackage{amsmath,amsfonts}
\usepackage{algorithmic}
\usepackage{algorithm}
\usepackage{array}
\usepackage[caption=false,font=normalsize,labelfont=sf,textfont=sf]{subfig}
\usepackage{textcomp}
\usepackage{stfloats}
\usepackage{url}
\usepackage{verbatim}
\usepackage{graphicx}
\usepackage{booktabs}
\usepackage{makecell}
\usepackage{amssymb}
\usepackage{multirow}
\usepackage{amsmath}
\usepackage{adjustbox}
\usepackage{tikz}
\usepackage[colorlinks,
           linkcolor=blue,
           anchorcolor=blue,
           citecolor=blue]{hyperref}
\begin{document}

%\title{A Sample Article Using IEEEtran.cls\\ for IEEE Journals and Transactions}
\title{TriCons-Pose: Triangle-Invariant Geometric Consistency Learning for Category-Level Object Pose Estimation}

\author{Zuzhi~Yang, Bingtao~Ma, Mounir~Kaaniche,~\IEEEmembership{Senior Member,~IEEE}, Ziwei~Li, Zhiming~Cheng, Zhidong~Zhao, Chenggang~Yan, Shuai~Wang
        % <-this % stops a space
\thanks{Zuzhi~Yang, Bingtao~Ma, Zhidong~Zhao, and Shuai~Wang are with the School of Cyberspace, Hangzhou Dianzi University, Hangzhou 310018, China (e-mail: 242270060@hdu.edu.cn, mabingtao93@gmail.com, zhaozd@hdu.edu.cn, shuaiwang.tai@gmail.com).}%% <-this % stops a space

\thanks{Zhiming~Cheng and Chenggang~Yan are with the School of Communication Engineering, Hangzhou Dianzi University, Hangzhou 310018, China (e-mail: czming@hdu.edu.cn, cgyan@hdu.edu.cn).}

\thanks{Mounir~Kaaniche is with Université Sorbonne Paris Nord, L2TI, UR 3043, F-93430, Villetaneuse, France, and Université Paris-Saclay, CentraleSupélec, CVN, 91190 Gif-sur-Yvette, France (e-mail: mounir.kaaniche@univ-paris13.fr).}

\thanks{Ziwei~Li is  with the School of Computer Science and Technology, University of Science and Technology of China, Hefei, China. (e-mail: ziwei.li@kaust.edu.sa).}
}

\markboth{Journal of \LaTeX\ Class Files,~Vol.~14, No.~8, August~2021}%
{Shell \MakeLowercase{\textit{et al.}}: A Sample Article Using IEEEtran.cls for IEEE Journals}


\maketitle
\raggedbottom

\begin{abstract}
Category-level object pose estimation is a crucial yet challenging task in both academia and industry, and has achieved remarkable success by leveraging keypoint-based correspondence paradigms. 
However, most existing methods increasingly rely on stronger feature learning while overlooking whether the established correspondences are geometrically stable across diverse perturbations. This often results in fragile pose recovery under intra-class shape variations and occlusions.
To tackle this challenge, we develop a novel Triangle-Invariant Geometric Consistency Learning for Category-Level Object Pose Estimation (TriCons-Pose) to anchor stable keypoints and aggregate pose-invariant cues, yielding reliable canonical mapping and accurate pose estimation. Specifically, a Structure-Consistent Keypoint Detector (SCKD) is designed to identify robust keypoints by enforcing cross-view structural consistency via normalized pairwise distance matching. Moreover, we propose a Pose-Invariant Geometric Aggregator (PIGA) to augment keypoint representations by injecting triangle-based pose-invariant descriptors into a local-to-global attention mechanism. The proposed framework is optimized using standard objective functions while incorporating an additional geometry consistency loss.   Extensive experiments on REAL275, CAMERA25, and HouseCat6D datasets demonstrate the effectiveness of the proposed approach.
\end{abstract}


\section{Introduction}
% 第一部分：姿态估计的背景和主要任务
\IEEEPARstart{O}{bject} pose estimation has received considerable attention in both academia and industry because of its broad applicability in robotic grasping \cite{cao20236impose}, autonomous driving~\cite{papaioannidis2022fast}, augmented reality \cite{manawadu2024advancing}, and 3D scene understanding \cite{nie2020total3dunderstanding}. 
% In particular, related tasks such as fast CNN-based human pose estimation \cite{papaioannidis2022fast} have demonstrated the critical importance of real-time perception for the safety and efficiency of autonomous systems.
As a fundamental problem in 3D vision, 6D object pose estimation seeks to determine the 3D rotation, 3D translation of an object, which are essential for geometric perception and subsequent physical interaction.
%第二部分，简单介绍一下实例级位姿估计并引出类别级位姿估计
With the support of instance-specific CAD models, instance-level object pose estimation~\cite{he2020pvn3d} has achieved remarkable success, yielding accurate pose predictions for seen object instances. However, its dependence on predefined object models restricts applicability to objects without corresponding CAD models and limits generalization to unseen instances. To overcome these limitations, category-level object pose estimation \cite{wang2019normalized,chen2020learning} seeks to estimate the pose and size of unseen objects from categories observed during training without requiring CAD models during inference. Despite recent progress, this task remains highly challenging due to intra-class shape variation, occlusion, and sensor noise.

%第三部分：介绍现有类别级位姿估计的的分类：
%（1）依靠形状先验的方法，存在问题
%（2）无形状先验的方法，面临的挑战
Existing category-level methods can generally be divided into prior-based and prior-free approaches. Prior-based methods \cite{tian2020shape,chen2021sgpa} learn category-level shape priors from CAD models and deform them to fit unseen instances. While effective in modeling category shape, they usually require large-scale 3D model collections and often generalize poorly to instances whose shapes deviate significantly from the learned mean prior. In contrast, prior-free methods \cite{chen2020learning,lin2021dualposenet} typically adopt normalized object coordinate space (NOCS)~\cite{wang2019normalized} or related canonical representations and estimate pose via dense correspondences between observations and canonical coordinates. Although more flexible, these methods often struggle with severe intra-class variation, under which dense correspondence learning becomes ambiguous and unstable.

%第四部分：引出AGpose这类，基于关键点的位姿估计范式，肯定其优点
To alleviate the difficulty of dense correspondence matching, recent studies have explored the keypoint-based correspondence paradigm \cite{lin2024instance,yu2025keypose,yang2025mk}. These methods detect a sparse set of instance-adaptive keypoints in camera space, and predict their correspondences in a canonical space to recover pose and size  \cite{wang2019normalized}. Compared with dense prediction, this paradigm provides compact and geometrically meaningful anchors, reduces interference from irrelevant local details, and is better suited for handling intra-class variation. Therefore, it has become an appealing direction for category-level pose estimation.
%第五部分：AGpose这类基于关键点的方法，存在的问题
However, existing methods mainly focus on stronger feature learning, better local-to-global interaction, or enhanced robustness to noise and domain bias \cite{yang2025mk,lin2025cleanpose,chen2020category,li2023sd}. We argue that the main bottleneck lies more fundamentally in the geometric instability of keypoint correspondences. Specifically, predicted keypoints may drift or change their structural configuration across viewpoint changes and perturbations, making them unreliable as correspondence anchors. Meanwhile, the geometric cues used for feature aggregation are often pose-sensitive. When pose-variant geometric encodings are adopted \cite{deng20193d}, the semantics of keypoint features may drift under SE(3) transformations, further degrading canonical correspondence prediction. These instabilities ultimately impair pose and size estimation.
%第六部分：针对上面的问题的讨论，引出我们的方法
This observation raises a fundamental problem: how can keypoint correspondence learning be stabilized under simultaneous intra-class shape variation, viewpoint changes, and input perturbations? We assume that a robust keypoint-based pose estimator should satisfy two requirements: stable keypoint anchors and pose-invariant geometric aggregation cues. The former ensures structurally consistent anchors across views and perturbations, while the latter enables transformation-stable and discriminative keypoint representations.

% 第七部分：具体介绍一下我们的方法
To address the above problems,  we propose Triangle-Invariant Geometric Consistency Learning for Category-Level Object Pose Estimation (TriCons-Pose), which is a simple yet effective framework that stabilizes correspondence learning in two aspects. Specifically, to obtain stable anchors, we design a \emph{Structure-Consistent Keypoint Detector} (SCKD). It enforces cross-view structural consistency by extracting keypoints from two randomly augmented views of the same instance and matching their normalized pairwise distance matrices. This introduces an order-invariant regularization that stabilizes the intrinsic geometry of the keypoint set. 
Moreover, to stabilize geometric cues in feature aggregation, we propose a \emph{Pose-Invariant Geometric Aggregator} (PIGA). It injects triangle-based pose-invariant geometric descriptors into local cross-attention with neighborhood points and global self-attention among keypoints. In this way, PIGA provides transformation-stable geometric biases and enhances the stability and discriminativeness of keypoint representations under SE(3) variations. By jointly stabilizing correspondence anchors and aggregation cues, our method substantially improves correspondence reliability and leads to robust category-level pose estimation.
% 第八部分：总结本文的四个贡献
Extensive experiments on three benchmarks confirm the effectiveness of our method. In particular, our method achieves an accuracy of 62\% under the $5^\circ$2cm metric and consistently improves performance on other evaluation metrics, while showing strong robustness in challenging scenarios involving occlusion, clutter, and severe intra-class shape variation. The main contributions  are summarized as follows:
\begin{itemize}
    \item We revisit category-level pose estimation from the perspective of keypoint correspondence and identify geometric instability as a key limiting factor.
    \item We propose SCKD, which stabilizes keypoint anchors via cross-view structural consistency using normalized pairwise distance matching.
    \item We propose PIGA, which injects triangle-based pose-invariant geometric cues into local-to-global attention for robust geometric feature aggregation.
    \item Extensive experiments demonstrate that our method substantially improves category-level pose  estimation and achieves strong robustness in challenging scenarios.
\end{itemize}
%The remainder of this paper is organized as follows. Section~II provides an overview of the existing category-level pose estimation methods. Section III describes the proposed TriCons-Pose framework. Finally, the experimental results are shown and discussed in Section IV, and some conclusions are drawn in Section V.

\section{Related Work}
%第一部分：一个小的Overview

%第二部分：Prior-Based 方法的基本范式，它的发展与不足
\subsection{Prior-Based Methods}
Prior-based methods typically learn category-level shape priors from CAD models of seen objects and then use them to guide correspondence learning or pose estimation. Early representative methods such as SPD \cite{tian2020shape} and SGPA \cite{chen2021sgpa} employed category-level mean shapes or learned shape priors as anchors to support deformation and alignment for pose recovery. These methods improve pose estimation by introducing category-aware geometric priors and provide effective regularization when the observed instance is close to the learned shape prior.
Subsequent studies improved prior-based methods through stronger prior deformation and discrepancy modeling \cite{lin2022category}, \cite{liang2025sdpose}. Nevertheless, the performance of prior-based methods is still strongly constrained by the quality and representativeness of the learned priors. When a test instance deviates significantly from the mean category shape or suffers from severe occlusion and incomplete observations, prior-guided alignment often becomes unreliable.
%第三部分：Prior-Free 方法的基本范式，存在的挑战
\subsection{Prior-Free Methods}
Prior-free methods directly estimate object pose from observations without explicitly relying on learned category priors. The Normalized Object Coordinate Space (NOCS) \cite{wang2019normalized} established a standard benchmark for this line of research by learning dense correspondences between camera-space observations and a unified canonical space. Subsequent methods further explored canonical shape learning and correspondence-based pose recovery \cite{chen2020learning,lin2021dualposenet}. Besides geometric modeling, the mathematical parameterization of the pose space also plays a vital role. For example, multi-objective quaternion learning~\cite{papaioannidis20193d} has been investigated to improve the stability of 3D object pose estimation by balancing different learning objectives. Prior-free methods are generally more flexible than prior-based methods, since they avoid the dependence on explicit category priors and can be trained in an end-to-end manner.
However, this flexibility comes with its own challenges. \emph{In particular}, dense correspondence prediction becomes increasingly ambiguous when different instances within the same category exhibit highly diverse geometry and structure. To address this issue, more recent methods have focused on strengthening geometric modeling, enhancing feature interaction, improving pose fitting strategies, and enforcing consistency learning to establish more reliable correspondences under large intra-class geometric and structural variation \cite{lin2021dualposenet,chen2021fs,liu2024mh6d}. Nonetheless, establishing reliable correspondence under large category-level variation remains a central challenge for prior-free methods.

\subsection{Keypoint-Based Correspondence Methods}
%第四部分：keypoint-based方法，以及后续的衍生方法
To alleviate the difficulty of dense correspondence learning, sparse keypoint-based methods have emerged as a promising alternative for category-level pose estimation. Instead of predicting dense correspondences, these methods represent object structure with a sparse set of geometrically meaningful keypoints, which serve as compact intermediates between observations and canonical object space~\cite{lin2024instance,yu2025keypose}. AG-Pose~\cite{lin2024instance} introduces an instance-adaptive keypoint detection module to improve adaptability to category-level shape variation without relying on predefined keypoint templates. KeyPose~\cite{yu2025keypose} further enhances this paradigm with a Transformer-based framework for keypoint-feature interaction and structural modeling. MK-Pose~\cite{yang2025mk} strengthens correspondence learning by incorporating richer multimodal semantic guidance. These studies demonstrate that sparse keypoints provide an effective bridge between observed point clouds and canonical representations for pose recovery. Beyond keypoint prediction itself, recent studies have also explored robustness under noise, occlusion, clutter, and data bias. For example, CleanPose~\cite{lin2025cleanpose} improves generalization through causal learning and knowledge distillation, while Diff-COPE~\cite{tang2025diff} models pose uncertainty with a conditional diffusion process. Other methods have investigated stronger geometric representations and refined feature interaction to mitigate the effects of incomplete observations and distribution shifts~\cite{li2023sd}.

%第五部分：现存keypoint-based方法不足之处，再顺带引出我的方案
Despite these advances, explicit geometric consistency supervision for sparse keypoint correspondence remains underexplored. Existing keypoint-based methods mainly emphasize prediction quality, multimodal interaction, or robustness enhancement, while paying less attention to whether the predicted keypoints preserve stable structural relationships across viewpoint changes and perturbations. Moreover, the geometric cues used for feature aggregation are often pose-sensitive, which may reduce the stability of keypoint representations. In contrast, our method places greater emphasis on geometric consistency by explicitly stabilizing sparse keypoint anchors through structure-consistent supervision and enhancing feature learning with pose-invariant geometric aggregation cues.

\section{Proposed Method}
\subsection{Overview}

\begin{figure*}[t]
    \centering
    \includegraphics[width=\textwidth]{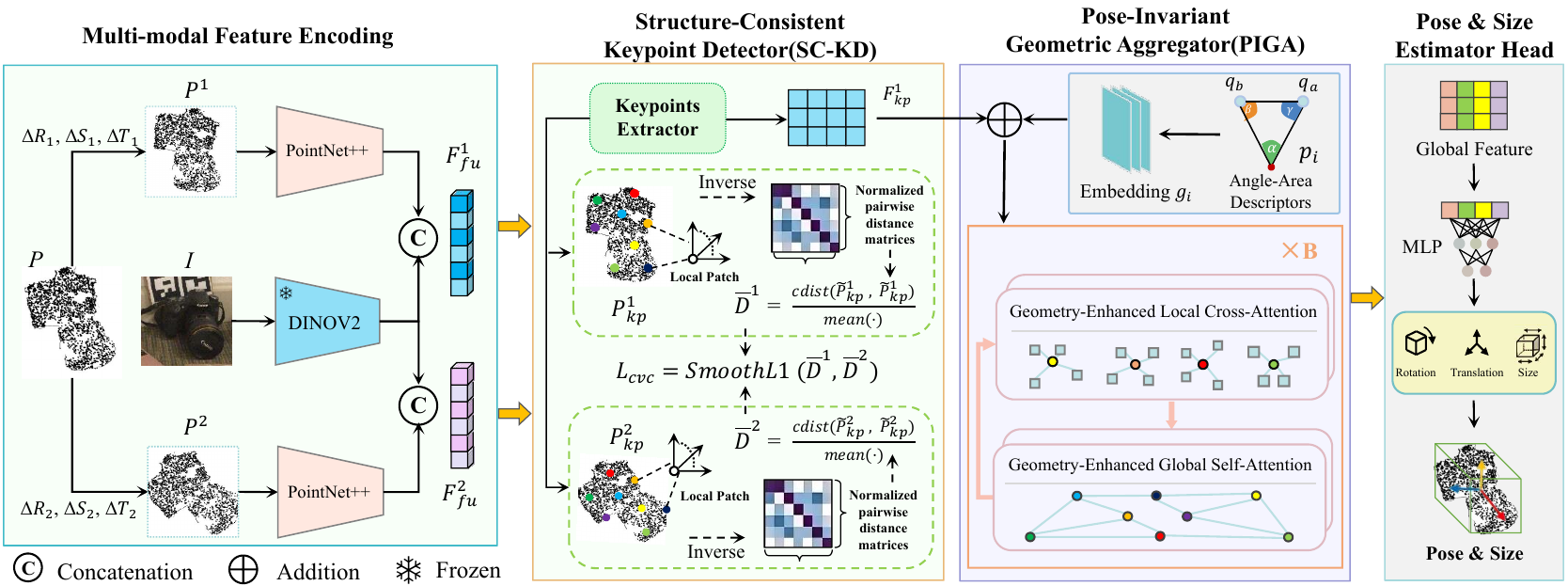}
    \caption{Overview of the proposed TriCons-Pose framework for category-level pose estimation. Given the cropped RGB image and the corresponding object point cloud, we extract image and geometric features and fuse them into multimodal per-point representations. The SCKD predicts instance-adaptive keypoints and stabilizes their spatial configuration by enforcing cross-view structural consistency through normalized pairwise distance matrices. The PIGA further enhances keypoint representations by introducing triangle-based pose-invariant geometric descriptors into a local-to-global attention mechanism. Finally, a lightweight pose head predicts the object rotation, translation, and anisotropic size from the aggregated keypoint features.}
    \label{fig:my_pipeline}
\end{figure*}

Given an RGB-D image, an offline Mask R-CNN \cite{kaiming2017mask} is employed to obtain the segmentation mask for object instance. For each segmented object, the mask is used to extract the cropped RGB image $I \in \mathbb{R}^{H \times W \times 3}$ and generate the corresponding object point cloud $P \in \mathbb{R}^{N \times 3}$ by back-projecting the segmented depth image with camera intrinsics followed by downsampling, where $N$ denotes the number of sampled points. Based on $I$ and $P$, our method aims to recover the 3D rotation $R \in SO(3)$, translation $t \in \mathbb{R}^3$, and anisotropic scale $s \in \mathbb{R}^3$ of the target object.

As illustrated in Fig.~\ref{fig:my_pipeline}, the proposed TriCons-Pose framework consists of four stages: Multi-modal Feature Encoding, Structure-Consistent Keypoint Detector (SCKD), Pose-Invariant Geometric Aggregator (PIGA), and Pose\&Size Estimator Head. Geometric and appearance features are first extracted from the point cloud $P$ and RGB image $I$ using PointNet++~\cite{fan2017point} and DINOv2~\cite{oquab2023dinov2}, respectively, and are subsequently fused into multimodal representations. Based on these fused features, SCKD is introduced to predict instance-adaptive keypoints. During training, structural consistency across two randomly augmented views is enforced by matching the normalized pairwise distance matrices of the predicted keypoints in a shared coordinate frame, thereby improving the stability of keypoint localization under perturbations. To enhance keypoint representations, PIGA is developed to incorporate triangle-based pose-invariant geometric cues into a local-to-global attention framework. Within this design, local triangle descriptors facilitate geometry-enhanced feature aggregation, while global self-attention among keypoints captures long-range structural dependencies. The refined keypoint features are then used to predict canonical coordinates and regress object rotation, translation, and anisotropic scale. By jointly stabilizing keypoint anchors and geometric feature aggregation, TriCons-Pose establishes more reliable correspondences for category-level pose estimation. The four proposed components are detailed in the following subsections.
%\vspace{-6pt}
\subsection{Multi-Modal Feature Encoding}
Given  the cropped RGB image 
$I$
 %the object point cloud 
%\in \mathbb{R}^{N \times 3}$ 
and the corresponding  point cloud $P$
%\in \mathbb{R}^{H \times W \times 3}$, 
we aim to construct multi-modal per-point representations for cross-view keypoint detection and subsequent correspondence learning. To this end, we generate two augmented point-cloud views, $P^{1}$ and $P^{2}$, from the original $P$ by applying two independently sampled rigid and scale transformations. It is formulated as:
\begin{equation}
P^{1}=\Delta S_1 \Delta R_1 P + \Delta T_1,\qquad
P^{2}=\Delta S_2 \Delta R_2 P + \Delta T_2,
\end{equation}
where $\Delta R\in SO(3)$, $\Delta T\in \mathbb{R}^{3}$, $\Delta S \in \mathbb{R}^{3}$ denote the randomly sampled rotation, translation, and anisotropic scale transformations. %Therefore, $P^{1}$ and $P^{2}$ denote two augmented views derived from the same object instance.

%For geometric encoding, the two augmented point clouds 
$P^{1}$ and $P^{2}$ are fed into a shared PointNet++ backbone~\cite{qi2017pointnet++} to extract point-wise geometric features $F_{p}^{1} \in \mathbb{R}^{N \times C_p}$ and $F_{p}^{2} \in \mathbb{R}^{N \times C_p}$.
%, where $C_p$ denotes the channel dimension of the geometric feature.
Meanwhile, we extract a dense feature map from image $I$ using a frozen DINOv2 backbone \cite{oquab2023dinov2}. The image features are then aligned to the object points through the sampling indices (e.g., \texttt{choose}) that record the correspondence between 3D points and image pixels, yielding per-point appearance features $F_{im} \in \mathbb{R}^{N \times C_{im}}$.
%, where $C_{im}$ is the appearance feature dimension. 

Since the two augmented views are generated by transforming the same point cloud in 3D space, the appearance features are shared across the two views and paired.
%with their corresponding geometric features. 
Finally, the geometric and appearance features are concatenated along the channel dimension to form the fused per-point representations:
\begin{equation}
F_{fu}^{1} = \mathrm{Cat}(F_{p}^{1}, F_{im}) \in \mathbb{R}^{N \times C},
\end{equation}
\begin{equation}
F_{fu}^{2} = \mathrm{Cat}(F_{p}^{2}, F_{im}) \in \mathbb{R}^{N \times C},
\end{equation}
where $C = C_p + C_{im}$.
The resulting fused features $F_{fu}^{1}$ and $F_{fu}^{2}$ integrate complementary geometric and appearance cues for two transformed views of the same object, and are subsequently fed into SCKD for cross-view keypoint learning and geometric consistency modeling.

\subsection{Structure-Consistent Keypoint Detector (SCKD)} 
%补充motivation 要解决的问题
Existing keypoint detectors lack explicit structural constraints, making the predicted keypoints prone to geometric drift across views and perturbations. To overcome this issue, the proposed Structure-Consistent Keypoint Detector (SCKD) predicts a set of instance-adaptive keypoints and explicitly regularizes the geometric structure of the keypoint set to remain stable under perturbations. 
%This module is designed to produce reliable correspondence anchors for category-level pose estimation.
SCKD produces reliable correspondence anchors for category-level pose estimation.

\subsubsection{Instance-Adaptive Keypoints Extractor}
\begin{figure}
    \centering
    \includegraphics[width=1\linewidth]{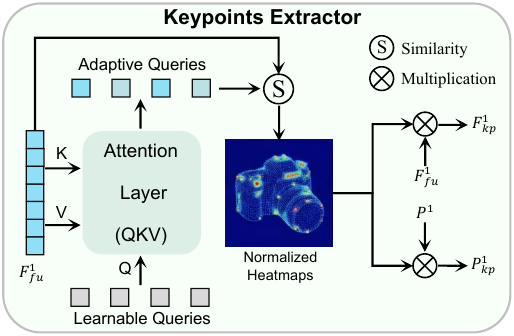}
    \caption{
    %Pipeline of the instance-adaptive keypoint detector. Learnable keypoint queries are first updated by attending to the fused point-wise features, yielding instance-adaptive query representations. The updated queries are then matched with the input features through similarity-based multiplication, and the resulting responses are normalized into heatmaps for keypoint localization.
    Pipeline of the instance-adaptive keypoint detector. Learnable keypoint queries are updated by attending to the fused point-wise features, yielding instance-adaptive query representations. The updated queries are then matched with the input features through similarity, and the resulting responses are normalized into heatmaps for keypoint localization.}
    \label{fig:keypoint}
\end{figure}
As illustrated in Fig. \ref{fig:keypoint}. Given the fused per-point features $F_{fu}^{1} \in \mathbb{R}^{N \times C}$ of the first augmented view, we first predict a set of instance-adaptive keypoint heatmaps for keypoint localization. Specifically, we introduce a set of learnable keypoint queries $Q^1 \in \mathbb{R}^{M \times C}$, where $M$ denotes the number of keypoints and each query is intended to capture one latent structural part of the object. These learnable queries interact with the fused point-wise features through an attention layer to produce instance-adaptive query embeddings:
\begin{equation}
Q_{ins}^1 = \mathrm{Attn}(Q^1, F_{fu}^{1}),
\end{equation}
%where $Q_{ins}^1 \in \mathbb{R}^{M \times C}$ denotes the updated query features conditioned on the current object instance. 
where $Q_{ins}^1 \in \mathbb{R}^{M \times C}$ denotes the updated query features for the current object instance. 
In this way, the initially shared learnable queries are dynamically adapted according to the input geometry and appearance, enabling flexible keypoint localization under large intra-category shape variations. 

Then, we compute similarity between  adapted queries and fused feature $F_{fu}^{1}$ to obtain the keypoint response map $S^{1} \in \mathbb{R}^{M \times N}$, whose $(i,j)$-th entry is defined as:
\begin{equation}
S^{1}_{ij} =
\frac{\langle Q^{1}_{i}, F_{fu,j}^{1} \rangle}
{\|Q^{1}_{i}\|_{2}\,\|F_{fu,j}^{1}\|_{2} + \epsilon},
\end{equation}
where $Q^{1}_{i}$ denotes the $i$-th adapted query, $F_{fu,j}^{1}$ denotes the fused feature of the $j$-th point, and $\epsilon$ is a small constant for numerical stability. 

Following this cosine-similarity matching, the response map is normalized along the point dimension to produce the keypoint heatmap $H^{1} = \mathrm{softmax}\!\left(S^{1}/\tau\right)$, where $\tau$ is a temperature factor controlling the sharpness of the heatmap distribution. Based on the predicted heatmap, the camera-space keypoint coordinates and the corresponding keypoint features are obtained by weighted aggregation over the input point cloud and fused point-wise features: $P_{kp}^{1} = H^{1} P^{1}$, $F_{kp}^{1} = H^{1} F_{fu}^{1}$, where $P_{kp}^{1} \in \mathbb{R}^{M \times 3}$ and $F_{kp}^{1} \in \mathbb{R}^{M \times C}$. 

Since each row of $H^{1}$ is normalized across all input points, each predicted keypoint is represented as a differentiable weighted combination of the entire point cloud. This formulation allows the detector to generate instance-adaptive keypoints while preserving stable feature support for subsequent geometric aggregation. The same keypoint extraction process is applied to the second augmented view $P^2$, yielding $H^{2}$ and $P_{kp}^{2}$, which are used together with the first-view predictions for cross-view structural consistency learning.

\subsubsection{Cross-View Structural Consistency}
Although the instance-adaptive keypoint extractor provides flexibility for representing different object instances, the predicted keypoints may still exhibit structural drift across transformed views of the same object. Such drift weakens the reliability of the detected keypoints as correspondence anchors. To alleviate this issue, we explicitly enforce cross-view structural consistency on two augmented views during training.

\begin{figure}
    \centering
    \includegraphics[width=1\linewidth]{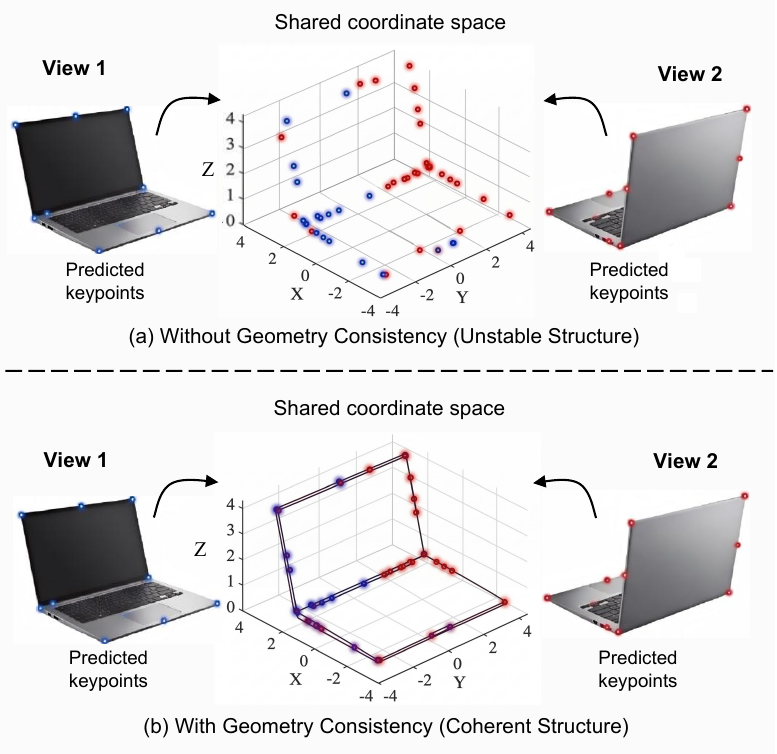}
    \caption{
    %Two augmented views of the same object are used to predict keypoints, which are then transformed into a shared coordinate space for structural comparison. (a) Without the geometry consistency constraint, the predicted keypoints from the two views (blue and red) exhibit noticeable discrepancies in their spatial arrangement, leading to an unstable geometric structure. (b) With the proposed geometry consistency loss applied during training, the predicted keypoints from different views become well aligned and preserve a coherent spatial configuration. The connecting edges illustrate the geometric relationships among keypoints, highlighting that the proposed loss effectively enforces structural consistency across views and stabilizes keypoint learning.
    Demonstration of the effectiveness of  cross-view structural consistency. (a) Without the structural consistency constraint, the predicted keypoints from the two views (blue and red) exhibit noticeable discrepancies in their spatial arrangement, leading to an unstable geometric structure. (b) With the proposed structural consistency loss applied during training, the predicted keypoints from different views become well aligned and preserve a coherent spatial configuration. The connecting edges illustrate the geometric relationships among keypoints, highlighting that the proposed loss effectively enforces structural consistency across views and stabilizes keypoint learning.}
    \label{fig:consistency}
\end{figure}

% Given the two augmented views $P^{1}$ and $P^{2}$ generated from the same original point cloud, the shared keypoint detector predicts two sets $P_{kp}^{1}, P_{kp}^{2} \in \mathbb{R}^{M \times 3}$. 
The keypoint sets $P_{kp}^1$ and $P_{kp}^2$ are mapped back to a shared reference frame by applying the inverse transformations corresponding to the point clouds augmentations, as illustrated in Fig. \ref{fig:consistency}. Specifically, the predicted keypoints are transformed back to the original coordinate system via:
\begin{equation}
\begin{aligned}
\widetilde{P}_{kp}^{1} &= \Delta R_1^{-1}\Delta S_1^{-1}(P_{kp}^{1}-\Delta T_1), \\
\widetilde{P}_{kp}^{2} &= \Delta R_2^{-1}\Delta S_2^{-1}(P_{kp}^{2}-\Delta T_2),
\end{aligned}
\end{equation}
where $\widetilde{P}_{kp}^{1}, \widetilde{P}_{kp}^{2} \in \mathbb{R}^{M \times 3}$ denote the aligned keypoint sets in the shared reference frame. Since the detected keypoints form an unordered set, explicit one-to-one matching between the two views is avoided. Instead, we adopt an order-invariant structural representation based on pairwise distance matrices:
\begin{equation}
D^{v} = \mathrm{cdist}\!(\widetilde{P}_{kp}^{v}, \widetilde{P}_{kp}^{v}), \qquad v \in \{1,2\},
\end{equation}
where $D^{v} \in \mathbb{R}^{M \times M}$ encodes the pairwise Euclidean distances among all predicted keypoints in view $v$.

To remove the influence of absolute scale, each distance matrix is normalized by its mean value:
\begin{equation}
\overline{D}^{v} =
\frac{D^{v}}
{\mathrm{mean}(D^{v}) + \epsilon}, \qquad v \in \{1,2\},
\end{equation}
where $\epsilon$ is a small constant to avoid numerical instability. The cross-view structural consistency loss is then defined as:
\begin{equation}
\mathcal{L}_{cvc}
=
\mathrm{SmoothL1}
(\overline{D}^{1}, \overline{D}^{2}).
\end{equation}
%We choose the Smooth L1 loss for cross-view structural consistency. 
The keypoint distance discrepancies between two augmented views may occasionally contain large deviations caused by augmentation and imperfect keypoint localization. In this case, Smooth L1 %is preferred as it %
is less sensitive to such outliers than L2, while preserving smooth optimization  near zero unlike standard L1~\cite{girshick2015fast}. This loss encourages the intrinsic geometric structure of the detected keypoints to remain invariant across perturbations, thereby producing more stable correspondence anchors. %without requiring explicit keypoint matching.

\subsection{Pose-Invariant Geometric Aggregator (PIGA)}
While adaptive keypoints provide compact correspondence anchors, their effectiveness largely depends on how geometric information is aggregated into keypoint features. Existing aggregation schemes typically rely on relative coordinates or spatial offsets \cite{jiang2024category}, which are inherently pose-sensitive and may lead to feature inconsistency under SE(3) transformations. To address this limitation, we propose the Pose-Invariant Geometric Aggregator (PIGA), which reformulates feature aggregation by introducing a \emph{pose-invariant geometric bias} derived from triangle-based invariants. Instead of encoding geometry through pose-variant spatial offsets, PIGA takes the predicted keypoint set $P_{kp}^{1}$ and its initial features $F_{kp}^{1}$ from SCKD as input and refines them using pose-invariant descriptors. 
% ------------------------------------------------
\subsubsection{Triangle-Based Pose-Invariant Geometry}
For each keypoint $p_{i} \in P_{kp}^{1}$, we collect its $K$ nearest neighboring points, denoted as $\{q_{ij}\}_{j=1}^{K}$. For every unordered neighbor pair $(q_{a}, q_{b})$, we construct a local triangle $\triangle(p_{i}, q_{ia}, q_{ib})$.
Each keypoint is associated with $N_t = K(K-1)/2$ local triangles. 
For each triangle, we compute its three inner angles: $(\alpha_{i,ab},\beta_{i,ab},\gamma_{i,ab}) = \Psi(p_i,q_a,q_b)$, where $\Psi(\cdot)$ denotes the angle computation operator, and $\alpha_{i,ab},\beta_{i,ab},\gamma_{i,ab}$ are the three inner angles of the triangle %formed by $p_i$, $q_a$, and $q_b$. 
Since triangle angles are invariant to rigid transformations, they provide stable local geometric descriptors under pose changes.

\begin{figure*}[t]
    \centering
    \includegraphics[width=\textwidth]{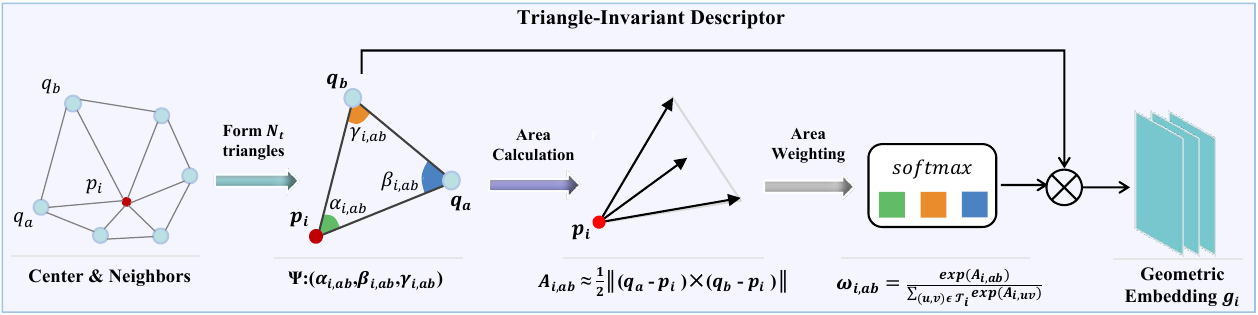}
    \caption{Triangle-invariant descriptor computation. For each center keypoint \(p_i\), its \(K\)-nearest neighboring points are selected to form $N_t$ local  triangles \(\triangle(p_i, q_a, q_b)\). The three interior angles \((\alpha_{i,ab}, \beta_{i,ab}, \gamma_{i,ab})\) are computed as rigid-transformation-invariant geometric descriptors. The triangle area \(A_{i,ab}\) is then estimated and normalized via a softmax function to generate the adaptive weight \(w_{i,ab}\). The weighted angle descriptor is finally encoded by an MLP to produce the local geometric embedding \(g_{i}\).}
    \label{fig:triangle}
\end{figure*}

To further emphasize geometrically salient local structures, we compute the area of each triangle as:
\begin{equation}
A_{i,ab} =
\frac{1}{2}
\left\|
(q_a - p_i) \times (q_b - p_i)
\right\|,
\end{equation}
where $A_{i,ab}$ is the area of triangle $\triangle(p_i,q_a,q_b)$. The triangle areas are normalized by a softmax function over $N_t$ triangles: %associated with the same keypoint:
\begin{equation}
w_{i,ab}
=
\frac{\exp(A_{i,ab})}
{\sum_{(u,v)\in \mathcal{T}_i}\exp(A_{i,uv})},
\end{equation}
where $\mathcal{T}_i$ denotes the set of all local triangles constructed around keypoint $p_i$. In this way, triangles with relatively larger areas receive higher weights and contribute more to the geometric representation.
The final weighted angle descriptor 
of each triangle 
is denoted as:
\begin{equation}
\phi_{i,ab}=w_{i,ab}\,[\alpha_{i,ab},\beta_{i,ab},\gamma_{i,ab}],
\end{equation}
where $\phi_{i,ab} \in \mathbb{R}^{3}$. These weighted angle descriptors are then encoded by an MLP to produce triangle-level pose-invariant geometric embeddings, and the overall computation process is illustrated in Fig. \ref{fig:triangle}.

% ------------------------------------------------
\subsubsection{Geometry-Enhanced Local Cross-Attention}
Let $F_{i} \in \mathbb{R}^{C}$ be the feature vector of keypoint $p_{i}$, which corresponds to the $i$-th row of the feature matrix $F_{kp}^{1}$, and let $\{F_{i,j}\}_{j=1}^{K}$ denote the features of its $K$ neighboring points, where each neighboring feature $F_{i,j} \in \mathbb{R}^{C}$. For keypoint $p_i$, the triangle-based invariant descriptors of all local triangles are first encoded by an MLP and then averaged over the triangle dimension to obtain a pose-invariant
%keypoint-level 
geometric embedding:
\begin{equation}
g_i = \frac{1}{N_t}\sum_{(a,b)\in \mathcal{T}_i} \mathrm{MLP}(\phi_{i,ab}) \in \mathbb{R}^{C},
\end{equation}
where $\mathcal{T}_i$ denotes the set of all local triangles constructed around keypoint $p_i$, and $N_t = |\mathcal{T}_i|$.

Instead of spatial offsets \cite{zhao2021point}, PIGA injects $g_i$ into neighboring point features before local cross-attention. Specifically, the geometry-enhanced neighboring features are defined as $\widehat{F}_{i,j} = F_{i,j} + g_i$, where $g_i$ is broadcast along the neighbor dimension. Queries, keys, and values are then computed as:
\begin{equation}
Q_i = W_q F_i,\quad
K_{i,j} = W_k \widehat{F}_{i,j},\quad
V_{i,j} = W_v \widehat{F}_{i,j},
\end{equation}
where $W_q$, $W_k$, and $W_v$ are learnable projection matrices.

The local cross-attention is formulated as:
\begin{equation}
\mathrm{Attn}(i,j)
=
\mathrm{softmax}
\left(
\frac{Q_i K_{i,j}^{\top}}{\sqrt{d}}
\right),
\end{equation}
where softmax is applied over $K$ neighboring points of keypoint $p_i$, and $d$ is channel dimension used for attention normalization. The final local feature is then computed as:
\begin{equation}
\ F^{b}_{\ell} = F^{b-1}_{\ell} + \sum_{j=1}^{K} \mathrm{Attn}(i,j)\, V_{i,j},
\end{equation}
where $b \in \{1, \dots, B\}$ denotes the  $b$-th local cross-attention block,
%index for the current geometric interaction block within the PIGA module,
$F_{\ell}^b$ and $F_{\ell}^{b-1}$ are the aggregated keypoint features for blocks $b$ and $b-1$, respectively. %For the first block ($b=1$), 
$F_{\ell}^0$ is 
%set to the starting 
keypoint features $F_{kp}^{1}$.

In this way, pose-invariant geometric information is integrated into local neighborhood aggregation through feature enhancement, which improves robustness to pose variation and enhancing the consistency of keypoint representations.
% ------------------------------------------------
\subsubsection{Geometry-Enhanced Global Keypoint Interaction}
Beyond local aggregation, we propose to further capture the global structural dependencies among all keypoints. %In this respect, %We first construct a small local neighborhood in the keypoint space and 
In the keypoint space, we first compute triangle-based invariant descriptors in the same manner as above. Then, we perform self-attention across the entire set of keypoints. Each keypoint feature is updated by aggregating information from all keypoints through self-attention, thereby integrating global geometric context into its representation. This global self-attention is performed on the geometry-enhanced keypoint features, preserving pose-invariance while modeling long-range relationships. 
Here, it is worth noting that  
%the distinction between the two attention mechanisms: the local cross-attention aggregates features between keypoints and their neighboring points, whereas 
the global self-attention operates among keypoints themselves to capture long-range dependencies.

Specifically, we use  $K_{gl}$ nearest neighboring keypoints for each keypoint, which yields $N_t^{{gl}}$ local triangles for each keypoint.
%The resulting triangle descriptors are encoded and averaged to obtain a global geometric embedding for each keypoint
These triangle descriptors are encoded and averaged to obtain a global geometric embedding for each keypoint:
\begin{equation}
g_i^{gl} = \frac{1}{N_t^{gl}} \sum_{(a,b)\in \mathcal{T}_i^{gl}} \mathrm{MLP}(\phi_{i,ab}^{gl}) \in \mathbb{R}^{C},
\end{equation}
where $\mathcal{T}_i^{gl}$ denotes the set of local triangles constructed from the neighboring keypoints of $p_i$.

Let $F_{\ell}$ denote the keypoint features enhanced by the local cross-attention block.
The geometric embedding is then added to these features before self-attention: $\widetilde{F}_{kp} = F_{\ell} + G_{gl}$, where $G_{gl} \in \mathbb{R}^{M \times C}$ is formed by stacking the per-keypoint embeddings $\{g_i^{gl}\}_{i=1}^{M}$. Self-attention is then performed on the geometry-enhanced keypoint features:
\begin{equation}
F^b_{gl} = F^{b-1}_{gl} + \mathrm{SelfAttn}(\mathrm{LN}(\widetilde{F}_{kp})),
\end{equation}
where $F^b_{gl} \in \mathbb{R}^{M \times C}$ denotes the aggregated keypoint feature matrix at the $b$-th global self-attention block. For the first block ($b=1$), it is initialized as $F_{\ell}$, $\mathrm{SelfAttn}(\cdot)$ is the standard multi-head self-attention applied across all keypoints, and $\mathrm{LN}(\cdot)$ denotes layer normalization.  
%After $B$ geometric interaction blocks, the final keypoint features $F \in \mathbb{R}^{M \times C}$ are acquired, which integrate both local triangle-invariant cues and global structural dependencies, preserving pose-invariance while enhancing feature discriminativeness.
The final keypoint features $F\in\mathbb{R}^{M \times C}$ are obtained by $B$ geometric interaction blocks, which integrate local triangle-invariant cues and global structural dependencies to preserve pose-invariance and enhance feature discriminativeness.

\begin{table*}[t]
\centering
\caption{\textbf{Quantitative comparisons with state-of-the-art methods on the REAL275 dataset.} The best results are shown in \textbf{bold}. The symbol -- indicates that the result is unavailable or not reported.}
\label{tab:real275_comparison}
\small
\setlength{\tabcolsep}{8pt}
\renewcommand{\arraystretch}{1.25}
\begin{tabular}{c|c|cc|cccc}
\Xhline{1.2pt}
Methods & Shape prior & 3D IoU$_{50}$ & 3D IoU$_{75}$ & 5$^\circ$2cm & 5$^\circ$5cm & 10$^\circ$2cm & 10$^\circ$5cm \\
\hline
NOCS (CVPR'19) \cite{wang2019normalized}         & $\times$     & 78.0          & 30.1          & 7.2           & 10.0          & 13.8          & 25.2 \\
SPD (ECCV'20) \cite{tian2020shape}          & $\checkmark$     & 77.3          & 53.2          & 19.3          & 21.4          & 43.2          & 54.1 \\
SGPA (ICCV'21) \cite{chen2021sgpa}         & $\checkmark$     & 80.1          & 61.9          & 35.9          & 39.6          & 61.3          & 70.7 \\
GPV-Pose (CVPR'22) \cite{di2022gpv}    & $\times$     & 83.0          & 64.4          & 32.0          & 42.9          & 55.0          & 73.3 \\
DPDN (ECCV'22) \cite{lin2022category}         & $\checkmark$ & 83.4          & 76.0          & 46.0          & 50.7          & 70.4          & 78.4 \\
IST-Net (ICCV'23) \cite{liu2023net}      & $\times$     & 82.5          & 76.6          & 47.5          & 53.4          & 72.1          & 80.5 \\
Query6DoF (ICCV'23) \cite{wang2023query6dof}    & $\checkmark$ & 82.5          & 76.1          & 49.0          & 58.9          & 68.7          & 83.0 \\
VI-Net (ICCV'23) \cite{lin2023vi}      & $\times$     & --            & --            & 50.0          & 57.6          & 70.8          & 82.1 \\
GPT-COPE (TCSVT'24) \cite{zou2023gpt}     & $\times$    & 82.0          & 70.4          & 45.9          & 53.8          & 63.1          & 77.7 \\
CLIPose (TCSVT'24) \cite{lin2024clipose}     & $\times$     & 83.6          & 76.6            & 48.7          & 58.3          & 70.4          & 85.2 \\
SecondPose (CVPR'24) \cite{chen2024secondpose}   & $\times$     & --            & --            & 56.2          & 63.6          & 74.7          & 86.0 \\
AG-Pose (CVPR'24) \cite{lin2024instance}     & $\times$     & \textbf{84.1} & 80.1          & 57.0          & 64.6          & 75.1          & 84.7 \\
SpherePose (ICLR'25) \cite{ren2025learning}   & $\times$     & 84.0          & 79.0          & 58.2          & 67.4          & 76.2          & \textbf{88.2} \\
GCE-Pose (CVPR'25) \cite{li2025gce}    & $\checkmark$ & \textbf{84.1} & 79.8          & 57.0          & 65.1          & 75.6          & 86.3 \\
MK-Pose (IROS'25) \cite{yang2025mk}     & $\times$     & 84.0          & 80.3          & 60.8          & --            & 78.0          & 84.6 \\
Clean-Pose (ICCV'25) \cite{lin2025cleanpose}  & $\times$     & --            & --            & 61.5          & 67.4          & \textbf{78.3} & 86.2 \\
\hline
TriCons-Pose (Ours)    & $\times$     & \textbf{84.1} & \textbf{80.5}          & \textbf{62.0} & \textbf{68.5} & 77.9          & 87.0 \\
\Xhline{1.2pt}
%\vspace{-5pt}
\end{tabular}
\end{table*}

\subsection{Pose and Size Estimation}
\subsubsection{Pose and Size Regression}
After obtaining the final geometric-enhanced keypoint features $F$ from PIGA, we follow the pipeline of AG-Pose \cite{lin2024instance} to predict the keypoint-level NOCS coordinates. Specifically, each keypoint's NOCS coordinate $P^{\text{nocs}}_{kp} \in \mathbb{R}^{M\times 3}$ is predicted via a shared MLP: $P^{\text{nocs}}_{kp} = \text{MLP}(F)$. Then, we construct the pose regression feature by concatenating the keypoint positions in camera space $P_{kp}$, the geometric-enhanced feature $F$, and the predicted NOCS coordinates:
%\begin{equation}
 $   f_{\text{pose}} = \text{Cat}\big[P_{kp}, F, P^{\text{nocs}}_{kp}\big].$
%\end{equation}
Finally, the 6D pose and size are regressed via separate MLP heads:
\begin{equation}
    [ R, t, s ] = [\text{MLP}_{R}(f_{\text{pose}}), \; \text{MLP}_{t}(f_{\text{pose}}), \; \text{MLP}_{s}(f_{\text{pose}})],
\end{equation}
where $R \in \mathbb{R}^{3\times 3}$ is represented in the continuous 6D rotation representation~\cite{zhou2019continuity}, $t \in \mathbb{R}^{3}$ is the residual translation relative to the point cloud centroid, and $s \in \mathbb{R}^{3}$ denotes the object size along each axis. 

%By adopting this keypoint-level NOCS prediction and pose regression strategy, our method effectively establishes robust keypoint correspondences for unseen instances. 

\subsubsection{ Objective Functions}
%To stabilize category-level learning, 
We retain the standard  supervision loss from the baseline framework \cite{lin2024instance} to stabilize category-level learning, including pose regression loss $\mathcal{L}_{pose}$, NOCS regression loss $\mathcal{L}_{nocs}$,  Chamfe Distance loss $\mathcal{L}_{cd}$    reconstruction loss $\mathcal{L}_{rec}$ , and  the proposed cross-view structural consistency loss  $\mathcal{L}_{cvc}$. 
%Specifically,  denotes the pose regression loss,  denotes the NOCS regression loss, $\mathcal{L}_{cd}$ denotes the Chamfer Distance loss \cite{fan2017point}, $\mathcal{L}_{div}$ denotes the diversity regularization loss, $\mathcal{L}_{rec}$ denotes the reconstruction loss, and $\mathcal{L}_{cvc}$ denotes the proposed cross-view structural consistency loss. 
%These objectives serve as auxiliary training signals and are kept unchanged except for the addition of $\mathcal{L}_{cvc}$ in our framework. 
The final objective function is fourmulated as:

\begin{equation}
\begin{aligned}
\mathcal{L}_{t} =\;& \lambda_{pose}\mathcal{L}_{pose}
+ \lambda_{nocs}\mathcal{L}_{nocs}
+ \lambda_{cd}\mathcal{L}_{cd}\\
&+ \lambda_{div}\mathcal{L}_{div} 
+ \lambda_{rec}\mathcal{L}_{rec}
+ \lambda_{cvc}\mathcal{L}_{cvc},
\end{aligned}
\end{equation}
where $\lambda_{pose}$, $\lambda_{nocs}$, $\lambda_{cd}$, $\lambda_{div}$, $\lambda_{rec}$, and $\lambda_{cvc}$ are the corresponding balancing coefficients.

\section{Experiments}
\subsection{Experimental Setup}
\subsubsection{Datasets}
We evaluate our method on the standard category-level pose estimation benchmarks CAMERA25 and REAL275 under the NOCS \cite{wang2019normalized} setting, and further validate its generalization ability on the more challenging HouseCat6D~\cite{jung2024housecat6d} dataset. CAMERA25 is a large-scale synthetic dataset containing diverse object instances with accurate pose and size annotations, while REAL275 consists of real-world indoor scenes captured by RGB-D sensors and exhibits significant occlusion, clutter, and depth noise. HouseCat6D is a recent real-world benchmark for category-level 6D object pose estimation, featuring larger intra-category shape variation, severe occlusion, background clutter, and more complex scene layouts, making it particularly challenging for robust pose estimation. Following previous works, models are trained on synthetic data and evaluated on REAL275 and HouseCat6D to assess cross-domain generalization performance. We adopt the official dataset splits and preprocessing protocols to ensure fair comparison with prior methods.

\subsubsection{Evaluation Metrics}
%We follow the standard NOCS~\cite{wang2019normalized} evaluation protocol for category-level pose estimation. Specifically, 
Following  NOCS~\cite{wang2019normalized}, we use 3D Intersection-over-Union (3D IoU) and pose accuracy under different error thresholds as evaluation protocol. 3D IoU is reported as the mean average precision (mAP) at different IoU thresholds between the ground-truth and predicted 3D bounding boxes, which jointly evaluates 6D pose and object size. In our experiments, we report IoU50 and IoU75. For pose evaluation, we adopt the widely used $n^\circ m$ cm metric, which measures the mean average precision under given rotation and translation error thresholds. A pose prediction is considered correct if the rotation error is smaller than $n^\circ$ and the translation error is less than $m$ cm. %Following previous works, we report pose accuracy under four thresholds: $5^\circ 2\text{cm}$, $(5^\circ 5\text{cm})$, $(10^\circ 2\text{cm})$, and $(10^\circ 5\text{cm})$, so as to provide a comprehensive evaluation under both strict and relatively relaxed pose criteria.

\subsubsection{Implementation Details}
The input image is cropped according to the segmentation mask and resized to $192 \times 192$, and the input point cloud is downsampled to $N=1024$ points for each object. The number of predicted keypoints is set to $M=96$, and the temperature factor is set to $\tau=0.1$. 
%In PIGA, the local neighborhood size is set to $K=16$, while the global keypoint neighborhood size is set to $K_{gl}=3$. 
%The point and image feature dimensions are set to $C_p=128$ and $C_{im}=128$, respectively, yielding a fused feature dimension of $C=256$. 
The number of PIGA blocks is set to $B=6$. The network is optimized using ADAM with an initial learning rate of $1\times10^{-4}$ and a triangular2 cyclical learning rate schedule. Data augmentation includes random translation $\Delta t \sim U(-0.02,0.02)$, random scaling $\Delta s \sim U(-0.8,1.2)$, and random rotation along each axis with angles sampled from $U(0,20)$. The loss weights are set to $\lambda_{cd}=2.0$, $\lambda_{nocs}=2.0$, $\lambda_{rec}=15.0$, $\lambda_{div}=10.0$, $\lambda_{pose}=0.3$, and $\lambda_{cvc}=0.1$. All experiments are conducted on a single NVIDIA RTX 3090 GPU with a batch size of 24.

\begin{table*}[t]
\centering
\caption{\textbf{Quantitative comparisons with state-of-the-art methods on the CAMERA25 dataset.} The best results are shown in \textbf{bold}. The symbol -- indicates that the result is unavailable or not reported.}
\label{tab:camera25_comparison}
\small
\setlength{\tabcolsep}{8pt}
\renewcommand{\arraystretch}{1.25}
\begin{tabular}{c|c|cc|cccc}
\Xhline{1.2pt}
Methods & Shape prior & 3D IoU$_{50}$ & 3D IoU$_{75}$ & 5$^\circ$2cm & 5$^\circ$5cm & 10$^\circ$2cm & 10$^\circ$5cm \\
\hline
GPV-Pose (CVPR'22) \cite{di2022gpv}    & $\times$     & 93.4          & 88.3          & 72.1          & 79.1          & --            & 89.0 \\
IST-Net (ICCV'23) \cite{liu2023net}     & $\times$     & 93.7          & 90.8          & 71.3          & 79.9          & 79.4          & 89.9 \\
Query6DoF (ICCV'23) \cite{wang2023query6dof}  & $\checkmark$ & 91.9          & 88.1          & 78.0          & 83.1          & 83.9          & 90.0 \\
VI-Net (ICCV'23) \cite{lin2023vi}     & $\times$     & --            & --            & 74.1          & 81.4          & 79.3          & 87.3 \\
CLIPose (TCSVT'24) \cite{lin2024clipose}   & $\times$     & --            & --            & 74.8          & 82.2          & 82.0          & 91.2 \\
AG-Pose (CVPR'24) \cite{lin2024instance}    & $\times$     & 93.8          & 91.3          & 77.8          & 82.8          & 85.5          & 91.6 \\
SpherePose (ICLR'25) \cite{ren2025learning}  & $\times$     & \textbf{94.8} & 92.4          & 78.3          & \textbf{84.3} & 84.8          & 92.3 \\
MK-Pose (IROS'25) \cite{yang2025mk}     & $\times$     & 94.1          & 92.2          & 77.9          & --            & 86.1          & 91.7 \\
Clean-Pose (ICCV'25) \cite{lin2025cleanpose} & $\times$     & --            & --            & \textbf{80.3} & 84.2          & 87.7          & 92.7 \\
\hline
TriCons-Pose (Ours)   & $\times$     & 94.3          & \textbf{92.7} & \textbf{80.3} & \textbf{84.3} & \textbf{88.0} & \textbf{93.4} \\
\Xhline{1.2pt}
\end{tabular}
\end{table*}

\subsection{Comparison With State-of-the-Art Methods}
The  proposed framework is compared to several recent state-of-the-art methods on the three aforementioned datasets. %It is worth pointing out that, for fair comparison, we considered the baseline methods adopted for each dataset evaluation task as performed in most recent related works.
\subsubsection{Performance on REAL275 Dataset}
%We compare TriCons-Pose with recent state-of-the-art category-level pose estimation methods on the REAL275 dataset following the standard NOCS \cite{wang2019normalized} evaluation protocol. 
As shown in Table~\ref{tab:real275_comparison}, TriCons-Pose achieves competitive or superior performance on the REAL275 dataset across most evaluation metrics. In particular, our method obtains the best accuracy of 62.0\% under the strict $5^{\circ}2\text{cm}$ metric, outperforming recent methods such as Clean-Pose~\cite{lin2025cleanpose} and MK-Pose~\cite{yang2025mk}. Under the $5^{\circ}5\text{cm}$ metric, TriCons-Pose also achieves the best performance of 68.5\%, surpassing all previous approaches. These results indicate that the proposed method provides more accurate pose estimation under strict evaluation criteria, where both rotation and translation errors are tightly constrained.
For the relatively relaxed $10^{\circ}2\text{cm}$ and $10^{\circ}5\text{cm}$ metrics, TriCons-Pose achieves competitive results of 77.9\% and 87.0\%, respectively, which are comparable to the best-performing methods such as Clean-Pose~\cite{lin2025cleanpose} and SpherePose~\cite{ren2025learning}. In terms of 3D bounding box alignment, TriCons-Pose obtains 84.1\% and 80.5\% under the IoU50 and IoU75 metrics, respectively, demonstrating strong performance in object shape alignment and overall pose estimation. It is worth noting that some competing methods, such as DPDN~\cite{lin2022category}, Query6DoF~\cite{wang2023query6dof}, and GCE-Pose~\cite{li2025gce}, rely on explicit shape priors to guide pose estimation. In contrast, TriCons-Pose achieves strong performance without requiring any shape prior during training, showing the effectiveness of the proposed prior-free correspondence learning framework.

The superior performance of TriCons-Pose mainly stems from its geometry-consistency design. Different from prior-based methods that depend on learned category shape priors, TriCons-Pose directly improves the reliability of sparse correspondences by stabilizing the intrinsic structure of predicted keypoints through SCKD. Meanwhile, PIGA introduces triangle-based pose-invariant geometric cues into local-to-global feature aggregation, making keypoint representations more robust to viewpoint change. 

These two components jointly es lies in
the geometric instability of learned correspondences, caused
by inconsistent keypoint anchors and pose-sensitive feature
aggregation

reduce correspondence ambiguity and lead to more stable pose recovery. 
%which explains the clear improvements under strict metrics such as $5^{\circ}2\text{cm}$ and $5^{\circ}5\text{cm}$ on REAL275. Overall, the results demonstrate that TriCons-Pose consistently improves pose estimation accuracy and achieves state-of-the-art performance among prior-free category-level pose estimation methods.

\subsubsection{Performance on CAMERA25 Dataset}
We further evaluate TriCons-Pose on the CAMERA25 dataset following the standard NOCS \cite{wang2019normalized} evaluation protocol. As shown in Table \ref{tab:camera25_comparison}, TriCons-Pose achieves competitive or superior performance across most evaluation metrics. In particular, our method obtains 80.3\% accuracy under the $5^{\circ}2\text{cm}$ metric, which is the same as  Clean-Pose \cite{lin2025cleanpose} and higher than  previous approaches such as AG-Pose \cite{lin2024instance} (77.8\%) and SpherePose \cite{ren2025learning} (78.3\%). Under the $5^{\circ}5\text{cm}$ metric, TriCons-Pose achieves 84.3\%, which achieves the best  performance. For the $10^{\circ}2\text{cm}$ and $10^{\circ}5\text{cm}$ metrics, our method achieves 88.0\% and 93.4\%, respectively. In particular, TriCons-Pose achieves the highest accuracy under the $10^{\circ}5\text{cm}$ metric, demonstrating strong robustness under relatively relaxed pose thresholds. In terms of 3D IoU evaluation, TriCons-Pose achieves 92.7\% under  IoU75, which is  the best results compared with previous methods. These results indicate that the proposed method maintains strong performance in both pose estimation accuracy and object shape alignment. It is worth noting that some competing approaches rely on explicit shape priors, while TriCons-Pose does not require any shape prior during training.

\begin{figure*}[t]
    \centering
    \setlength{\tabcolsep}{1pt}
    \begin{tabular}{c@{\hspace{3pt}}ccccc}
        \adjustbox{valign=c}{\rotatebox[origin=c]{90}{AG-Pose}}
        & \adjustbox{valign=c}{\includegraphics[width=0.185\textwidth]{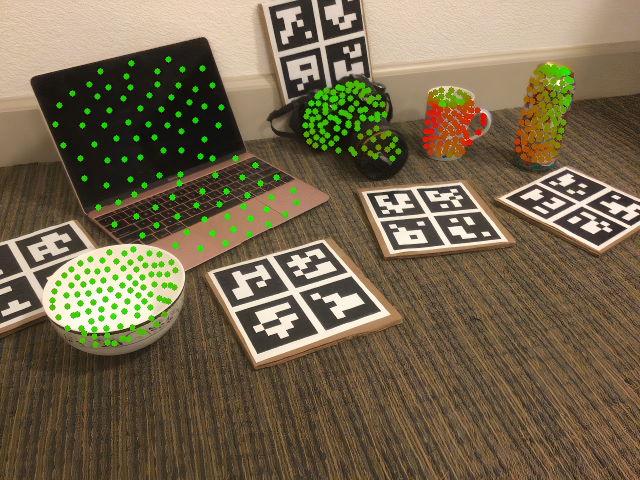}}
        & \adjustbox{valign=c}{\includegraphics[width=0.185\textwidth]{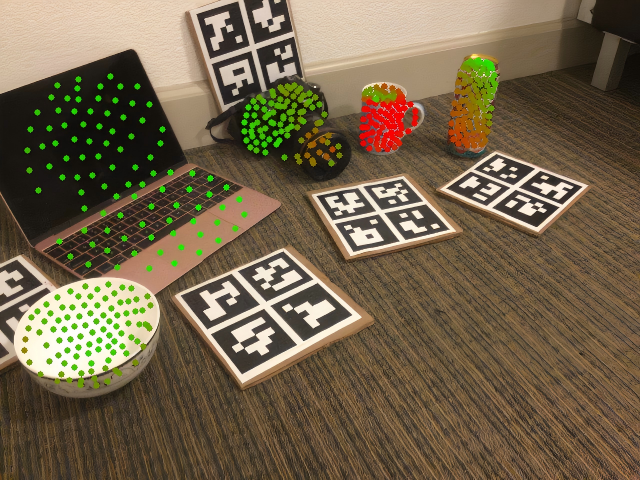}}
        & \adjustbox{valign=c}{\includegraphics[width=0.185\textwidth]{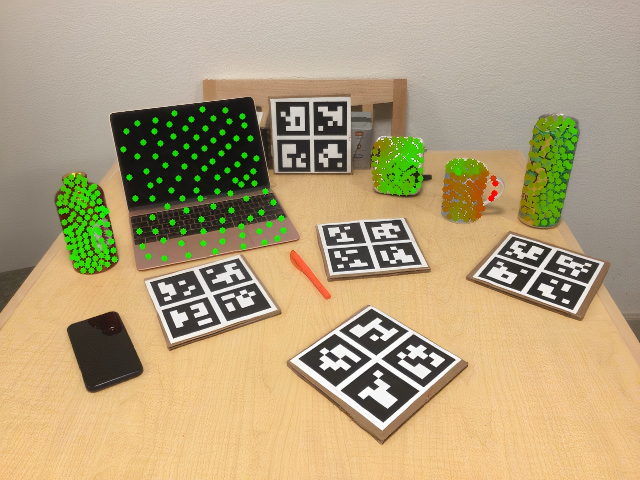}}
        & \adjustbox{valign=c}{\includegraphics[width=0.185\textwidth]{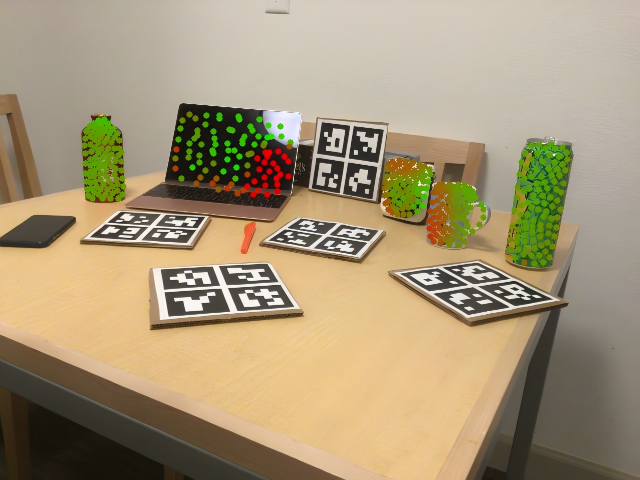}}
        & \adjustbox{valign=c}{\includegraphics[width=0.185\textwidth]{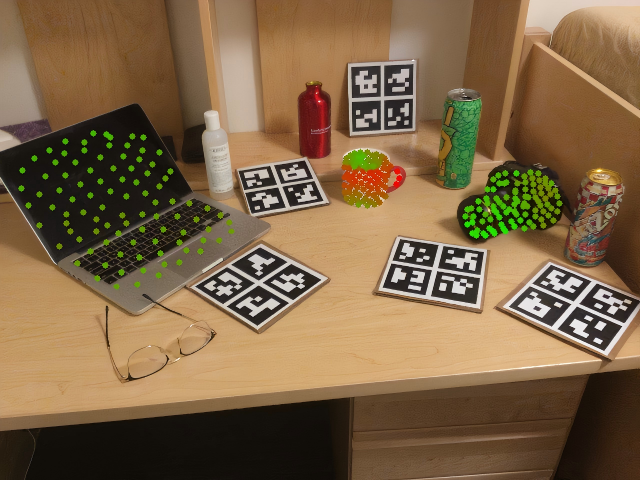}} \\

        \adjustbox{valign=c}{\rotatebox[origin=c]{90}{Ours}}
        & \adjustbox{valign=c}{\includegraphics[width=0.185\textwidth]{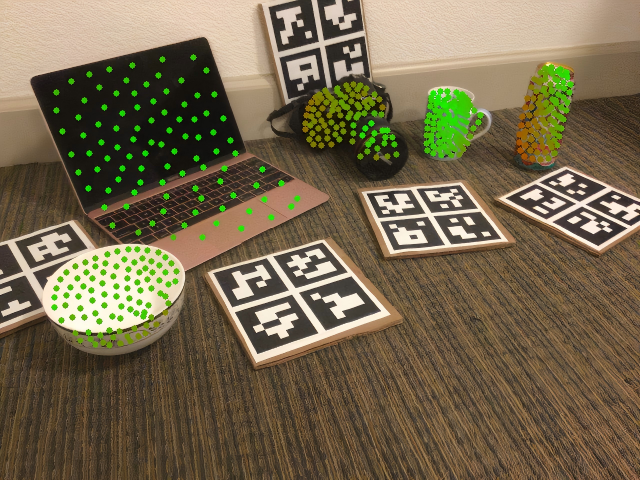}}
        & \adjustbox{valign=c}{\includegraphics[width=0.185\textwidth]{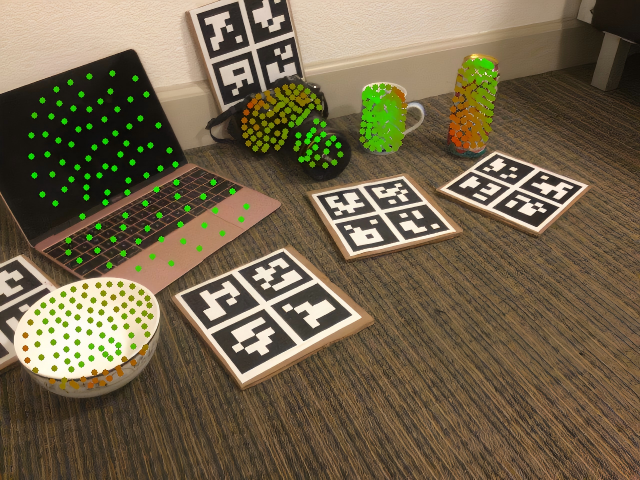}}
        & \adjustbox{valign=c}{\includegraphics[width=0.185\textwidth]{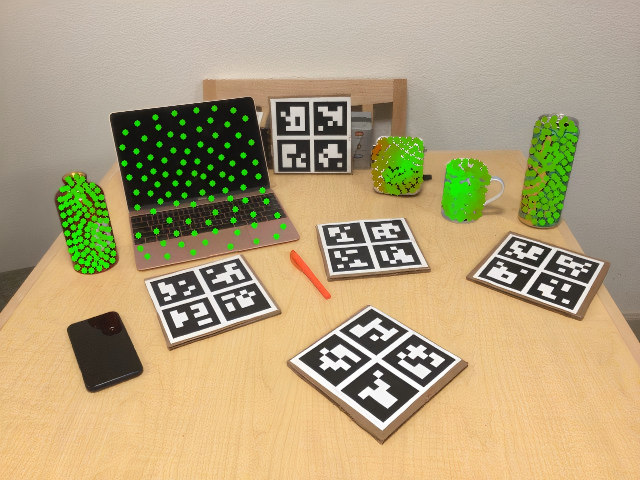}}
        & \adjustbox{valign=c}{\includegraphics[width=0.185\textwidth]{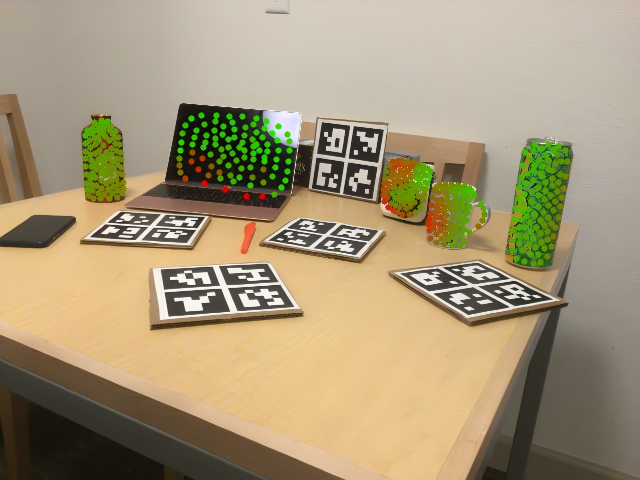}}
        & \adjustbox{valign=c}{\includegraphics[width=0.185\textwidth]{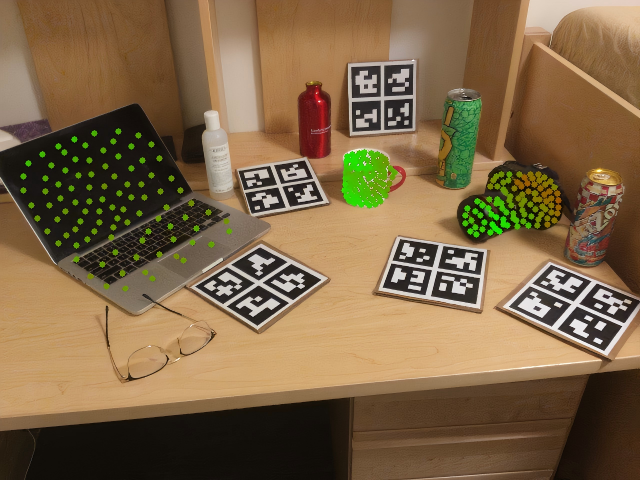}} \\

        & (a) & (b) & (c) & (d) & (e)
    \end{tabular}
    \caption{Visualization of predicted keypoints on the REAL275 dataset. We compare the keypoints generated by AG-Pose \cite{lin2024instance} and our TriCons-Pose on representative test samples. Keypoint colors indicate the NOCS correspondence errors, linearly mapped from green (error = 0) to red (error = 0.2). Compared with AG-Pose \cite{lin2024instance}, our method produces more accurate, uniformly distributed, and geometrically consistent keypoints, demonstrating the effectiveness of the proposed SCKD and PIGA modules.}
    \label{fig:keypoint_visualization}
\end{figure*}

\subsubsection{Performance on HouseCat6D Dataset}
We further evaluate TriCons-Pose on the HouseCat6D \cite{jung2024housecat6d} dataset, which is a more challenging benchmark for category-level 6D pose estimation due to its larger intra-category shape variation, severe occlusion, background clutter, and complex real-world scene layouts. As shown in Table \ref{tab:housecat6d_comparison}, TriCons-Pose achieves  state-of-the-art performance on HouseCat6D \cite{jung2024housecat6d} in terms . In particular, our method obtains the best accuracy of 25.1\% under the $5^{\circ}2\text{cm}$ metric, outperforming previous methods such as Clean-Pose~\cite{lin2025cleanpose} (22.4\%), AG-Pose~\cite{lin2024instance} (21.3\%), and SpherePose~\cite{ren2025learning} (19.3\%). Under the $5^{\circ}5\text{cm}$ metric, TriCons-Pose also achieves the best result of 26.4\%, surpassing Clean-Pose~\cite{lin2025cleanpose} (24.1\%) and other prior methods by a clear margin. These improvements under stricter pose thresholds indicate that the proposed method is able to produce more accurate and stable pose predictions in challenging real-world scenarios. Under relatively looser pose criteria, TriCons-Pose still maintains strong performance. Specifically, it achieves 53.5\% under the $10^{\circ}2\text{cm}$ metric, which is higher than Clean-Pose \cite{lin2025cleanpose} (51.6\%), AG-Pose~\cite{lin2024instance} (51.3\%), and SpherePose \cite{ren2025learning} (40.9\%). Under the $10^{\circ}5\text{cm}$ metric, TriCons-Pose reaches 56.5\%, achieving the best performance. These results demonstrate that our method remains robust even when evaluated under less strict but still practically meaningful pose thresholds. In terms of 3D IoU evaluation, TriCons-Pose achieves 76.6\% under the IoU50 metric, which is competitive with existing methods. Although Clean-Pose \cite{lin2025cleanpose} achieves a slightly higher IoU50 score of 79.8\%, TriCons-Pose consistently performs better on the stricter pose accuracy metrics.

\begin{table}[t]
\centering
\caption{\textbf{Quantitative comparisons with state-of-the-art methods on the HouseCat6D dataset.} The best results are shown in \textbf{bold}.}
\label{tab:housecat6d_comparison}
\footnotesize
\setlength{\tabcolsep}{3pt}
\renewcommand{\arraystretch}{1.2}
\resizebox{\columnwidth}{!}{%
\begin{tabular}{c|c|cccc}
\Xhline{1.2pt}
Methods & 3D IoU$_{50}$ & 5$^\circ$2cm & 5$^\circ$5cm & 10$^\circ$2cm & 10$^\circ$5cm \\
\hline
GPV-Pose (CVPR'22) \cite{di2022gpv}    & 50.7          & 3.2           & 4.6           & 17.8          & 22.7 \\
VI-Net (ICCV'23) \cite{lin2023vi}      & 56.4          & 8.4           & 10.3          & 20.5          & 29.1 \\
AG-Pose (CVPR'24) \cite{lin2024instance}     & 76.9          & 21.3          & 22.1          & 51.3          & 54.3 \\
SpherePose (ICLR'25) \cite{ren2025learning}  & 72.2          & 19.3          & 25.9          & 40.9          & 55.3 \\
Clean-Pose (ICCV'25) \cite{lin2025cleanpose} & \textbf{79.8} & 22.4          & 24.1          & 51.6          & \textbf{56.5} \\
\hline
TriCons-Pose (Ours)   & 76.6          & \textbf{25.1} & \textbf{26.4} & \textbf{53.5} & \textbf{56.5} \\
\Xhline{1.2pt}
\end{tabular}%
}
\end{table}

\subsubsection{Visualization of Predicted Keypoints}
Fig. \ref{fig:keypoint_visualization} presents a qualitative comparison of the keypoints predicted by TriCons-Pose and the baseline method AG-Pose \cite{lin2024instance} on representative samples from the REAL275 dataset. To better reveal the quality of keypoint correspondences, the predicted keypoints are color-coded according to their NOCS correspondence errors, which are linearly mapped from green (error = 0) to red (error = 0.2). In this visualization, greener keypoints indicate more accurate correspondences, while redder keypoints indicate larger deviations from the ground-truth NOCS locations. As shown in Fig. \ref{fig:keypoint_visualization}, the keypoints predicted by AG-Pose \cite{lin2024instance} exhibit noticeably larger correspondence errors and less stable spatial distributions, especially on geometrically complex regions, object boundaries, and fine structures. In contrast, our TriCons-Pose generates keypoints with significantly lower NOCS errors, as reflected by the predominance of green points across different object instances and viewpoints. Moreover, the  keypoints predicted by our method cover the object surface more uniformly and preserve more consistent geometric structures, which is crucial for robust category-level pose estimation on real-world data. In particular, under cluttered backgrounds, viewpoint changes, and partial occlusions commonly present in REAL275, AG-Pose \cite{lin2024instance} often fails to place keypoints on informative geometric parts and tends to produce scattered predictions with larger errors. In contrast, our method yields more compact and precise keypoints  that better align with the underlying object geometry. These qualitative results are consistent with the quantitative improvements reported above and further demonstrate that the superior pose estimation performance of TriCons-Pose stems from learning more accurate and structurally stable keypoint correspondences.

\subsubsection{Complexity Analysis}

%Following the notation defined above, where $N$, $M$, $K$, $K_{gl}$, $C$, and $B$ denote the number of input points, detected keypoints, local neighboring points, neighboring keypoints for global triangle construction, feature dimension, and aggregation blocks, respectively, 
We analyze the computational complexity of the keypoint detection and feature aggregation stages in TriCons-Pose and AG-Pose, excluding backbone feature extraction due to its architecture-dependent cost.

For AG-Pose, the instance-adaptive keypoint detector computes keypoint-query and point-feature interactions with complexity $\mathcal{O}(MNC)$. The Geometric-Aware Feature Aggregation (GAFA) module further aggregates local features from $K$ neighboring points and models global relationships among $M$ keypoints. Thus, with $B$ GAFA blocks, its inference complexity is given by
\begin{equation}
\mathcal{O}_{\mathrm{AG}} = \mathcal{O}\left(MC\left(N + B(K + M)\right)\right),
\end{equation}
where $N$ and $C$ denote the number of input points and feature dimension, respectively.
\begin{table}[t]
\centering
\caption{Effectiveness of different components on the real275 dataset.}
\label{tab:ablation_first}
\small
\setlength{\tabcolsep}{7pt}
\renewcommand{\arraystretch}{1.2}
\begin{tabular}{c|cccc}
\Xhline{1.2pt}
Methods & 5$^\circ$2cm & 5$^\circ$5cm & 10$^\circ$2cm & 10$^\circ$5cm \\
\hline
AG-Pose (Baseline)  & 57.0 & 64.6 & 75.1 & 84.7 \\
w/o PIGA            & 58.5 & 65.3 & 77.2 & 85.8 \\
w/o SCKD           & 60.5 & 67.0 & 77.6 & 86.7 \\
Full                & \textbf{62.0} & \textbf{68.5} & \textbf{77.9} & \textbf{87.0} \\
\Xhline{1.2pt}
\end{tabular}
\end{table}

TriCons-Pose keeps the same keypoint detection complexity of $\mathcal{O}(MNC)$. In each PIGA block, local triangle construction over $K$ neighboring points introduces a descriptor computation cost of $\mathcal{O}(MK^{2}C)$, local triangle-aware cross-attention cost of $\mathcal{O}(MKC)$, global keypoint self-attention cost of $\mathcal{O}(M^{2}C)$, and global triangle descriptors over $K_{gl}$ neighboring keypoints of cost $\mathcal{O}(MK_{gl}^{2}C)$. Therefore, with $B$ PIGA blocks, the overall inference complexity is:
\begin{equation}
\mathcal{O}_{\mathrm{TriCons}} = \mathcal{O}\left(MC\left(N + B(K^{2} + K + K_{gl}^{2} + M)\right)\right).
\end{equation}

The KNN search for neighborhood construction is shared by both methods and is omitted as an implementation-dependent operation. Compared with AG-Pose, TriCons-Pose mainly adds the local and global triangle descriptor terms, which correspond to $\mathcal{O}(MK^{2}C)$ and $\mathcal{O}(MK_{gl}^{2}C)$. 
%Since these computations are performed only on sparse keypoints and compact neighborhoods, the overhead remains moderate. 
In our implementation, $N=10
24$, $M=96$, $K=16$, $K_{gl}=3$, $C=256$, and $B=6$, where $M \ll N$, $K \ll N$, and $K_{gl} \ll M$, thus the additional geometric modeling cost is marginal.
Moreover, our method avoids using high-complexity manner such as dense attention cost of $\mathcal{O}(N^{2}C)$ to improve performance. 
%Moreover, our method avoids the dense attention cost of $\mathcal{O}(N^{2}C)$.
%Since these computations are performed only on sparse keypoints and compact neighborhoods, the overhead remains moderate. In our implementation, where M ≪ N, K ≪ N, and Kgl ≪ M. 
%As a result, the additional geometric modeling cost is marginal and avoids the dense attention complexity of O(N^{2}C).

%The cross-view structural consistency loss in SCKD is used only during training and thus does not affect inference complexity.
%It computes and matches normalized pairwise keypoint distance matrices with complexity $\mathcal{O}(M^{2})$, and thus does not affect inference complexity.
Overall, TriCons-Pose preserves the sparse keypoint-based paradigm of AG-Pose while introducing only compact triangle-invariant geometric overhead for more stable keypoint anchors and pose-invariant aggregation.

\subsection{Ablation Study}
\subsubsection{Effect of cross-view Structural Consistency and Triangle-Invariant Geometry}
To evaluate the contribution of the proposed cross-view structural consistency constraint and triangle-invariant geometric descriptors, we conduct an ablation study with four variants: AG-Pose (Baseline), w/o PIGA, w/o SCKD, and Full. 
%These variants differ in whether the cross-view structural consistency loss in SCKD and the triangle-invariant geometric descriptors in PIGA are employed. 
Table \ref{tab:ablation_first} presents the comparison results of these variants on the REAL275 dataset.
AG-Pose (Baseline) removes both the cross-view structural consistency loss and the triangle-based geometric descriptors.
The variant w/o PIGA remains the cross-view structural consistency loss while removing the triangle-invariant descriptors. 
The variant w/o SCKD  removes the  cross-view structural consistency loss  only. 
The results of w/o PIGA demonstrate that enforcing cross-view structural consistency improves model performance by stabilizing the spatial configuration of predicted keypoints and reducing keypoint drift under pose perturbations. 
Compared with AG-Pose (Baseline), w/o SCKD also achieves performance gains by incorporating triangle-invariant geometric descriptors into the feature aggregation module. This indicates that pose-invariant geometric cues provide more reliable structural information for keypoint feature learning. 
%When both the cross-view structural consistency loss and triangle-invariant descriptors are integrated (Full), 
Our Full model achieves the best performance, which demonstrates that the two components are complementary: the cross-view structural consistency constraint stabilizes keypoint anchors, while the triangle-based geometric descriptors improve geometry-enhanced feature aggregation. Their combination significantly enhances the reliability of learned correspondences and leads to more accurate pose estimation.

\begin{table}[t]
\centering
\caption{Ablation study of different geometry consistency losses
%We compare different losses for enforcing keypoint geometry consistency
on the real275 dataset. The best results are shown in \textbf{bold}.}
\label{tab:ablation_geom_loss}
\small
\setlength{\tabcolsep}{8pt}
\renewcommand{\arraystretch}{1.2}
\begin{tabular}{c|cccc}
\Xhline{1.2pt}
Loss type & 5$^\circ$2cm & 5$^\circ$5cm & 10$^\circ$2cm & 10$^\circ$5cm \\
\hline
Cosine & 60.1 & 66.1 & \textbf{78.9} & 86.9 \\
Chamfer & 60.2 & 67.1 & 78.1 & \textbf{87.0} \\
Smooth L1 (ours) & \textbf{62.0} & \textbf{68.5} & 77.9 & \textbf{87.0} \\
\Xhline{1.2pt}
\end{tabular}
\end{table}

\subsubsection{The choice of geometry consistency loss}
To evaluate the effectiveness of the proposed geometry consistency formulation, we replace the Smooth L1 loss  with other losses to enforce cross-view structural consistency~\cite{qiu2025leveraging}. Specifically, three variants are compared: (1) cosine similarity loss, (2) Chamfer distance~\cite{fan2017point}, and (3) the Smooth L1 loss~\cite{lin2022category} applied to the normalized pairwise distance matrices of keypoints. As shown in the table~\ref{tab:ablation_geom_loss}, different loss functions lead to different effects on pose estimation performance. Cosine loss achieves the best result under the $10^{\circ}2\text{cm}$ metric, indicating that enforcing directional similarity between the flattened structural representations can help preserve global geometric patterns under relatively relaxed rotation thresholds. However, its performance under stricter metrics, such as $5^{\circ}2\text{cm}$ and $5^{\circ}5\text{cm}$, is inferior to that of the proposed formulation. Chamfer distance~\cite{fan2017point} also provides competitive performance, especially under the $10^{\circ}5\text{cm}$ metric. Nevertheless, since Chamfer distance relies on nearest-neighbor matching, it may introduce ambiguous supervision when multiple keypoints share similar spatial distributions.

In contrast, the Smooth L1-based consistency loss achieves the best performance under the strict $5^{\circ}2\text{cm}$ and $5^{\circ}5\text{cm}$ metrics, with accuracies of 62.0\% and 68.5\%, respectively. It also maintains competitive performance under looser pose thresholds and achieves the best result under the $10^{\circ}5\text{cm}$ metric. We attribute this advantage to the robustness of the Smooth L1 penalty, which balances sensitivity to geometric deviations and robustness to outliers. 
By directly penalizing discrepancies between normalized pairwise distance matrices, the proposed formulation preserves the intrinsic spatial relationships among keypoints while avoiding instability caused by augmentation, imperfect keypoint localization, and partial occlusion. 
%These results demonstrate that the choice of loss function plays an important role in enforcing cross-view structural consistency. 
Among the evaluated alternatives, the proposed Smooth L1-based formulation provides the most stable and effective supervision for cross-view keypoint structure learning.

\subsubsection{Ablation study of Geometric Descriptor}
We conduct an ablation study on the Real275 dataset about the effect of different geometric descriptors in the keypoint feature enhancement module. Specifically, we compare four alternatives, including Relative, PPF \cite{drost2010model}, the pose-invariant distance and angle-based descriptor \cite{qin2023geotransformer}, and the Triangle descriptor \cite{slimani2024rocnet++} adopted in our implementation, while keeping all other network components and training settings unchanged. As reported in Table \ref{tab:ablation_geom_descriptor}, the Triangle descriptor achieves the best performance under all evaluation thresholds, yielding 62.0, 68.5, 77.9, and 87.0 under $5^{\circ}2\mathrm{cm}$, $5^{\circ}5\mathrm{cm}$, $10^{\circ}2\mathrm{cm}$, and $10^{\circ}5\mathrm{cm}$, respectively. In contrast, the Relative descriptor produces inferior results, indicating that directly using relative geometric quantities is insufficient for capturing discriminative local structures. PPF \cite{drost2010model} improves over Relative in some settings, suggesting the benefit of incorporating transformation-invariant geometric cues. The distance and angle-based descriptor \cite{qin2023geotransformer} further improves the performance, showing that jointly modeling distance and angular relations provides a stronger pose-invariant representation. Among all compared variants, the Triangle descriptor consistently performs the best. Compared with the second-best descriptor, it improves the registration accuracy by 3.1, 2.6, 1.1, and 0.7 points, respectively. These show that the Triangle descriptor has strong capability in encoding higher-order local geometric relationships between the center point and pairs of neighboring points. It can provides more discriminative structural cues than conventional pairwise formulations. These results suggest that introducing a more expressive geometric descriptor into the feature enhancement module is critical for improving local feature quality and registration performance.
\begin{table}[t]
\centering
\caption{Ablation study of different geometric descriptors on the Real275 dataset.
%We compare several geometric descriptors used in the aggregation module. 
The best results are shown in \textbf{bold}.}
\label{tab:ablation_geom_descriptor}
\small
\setlength{\tabcolsep}{5pt}
\renewcommand{\arraystretch}{1.2}
\begin{tabular}{c|cccc}
\Xhline{1.2pt}
Geometric descriptor & 5$^\circ$2cm & 5$^\circ$5cm & 10$^\circ$2cm & 10$^\circ$5cm \\
\hline
Relative \cite{shaw2018self} & 57.0 & 64.6 & 75.1 & 84.7 \\
PPF \cite{drost2010model}      & 57.4 & 65.8 & 73.9 & 85.9 \\
Dis-Ang \cite{qin2023geotransformer}  & 58.9   & 65.9   & 76.8   & 86.3   \\
Triangle-based descriptor & \textbf{62.0} & \textbf{68.5} & \textbf{77.9} & \textbf{87.0} \\
\Xhline{1.2pt}
\end{tabular}
\end{table}

\section{\textbf{Conclusion}}
In this work, we revisited category-level 6D pose estimation from the perspective of correspondence stability. The main limitation of the existing keypoint-based pipelines lies in the geometric instability of learned correspondences, caused by inconsistent keypoint anchors and pose-sensitive feature aggregation. To address this issue, we proposed a unified framework that stabilizes correspondence learning from two aspects. First, we designed Structure-Consistent Keypoint Detector (SCKD) that enforces cross-view structural consistency for stable keypoint prediction. Moreover, we developed Pose-Invariant Geometric Aggregator (PIGA), which introduces triangle-based invariant descriptor to guide feature aggregation. Together, these designs lead to more reliable canonical correspondences and improved pose estimation performance. Beyond performance improvements, our results suggest that robustness in category-level pose estimation may rely less on stronger regressors and more on learning geometry-consistent and transformation-stable correspondences. We believe that this novel design will further encourage future research toward principled geometric representations for robust 3D perception.

\bibliographystyle{IEEEtran}
\bibliography{reference.bib}

\begin{IEEEbiography}[{\includegraphics[width=1in,height=1.25in,clip,keepaspectratio]{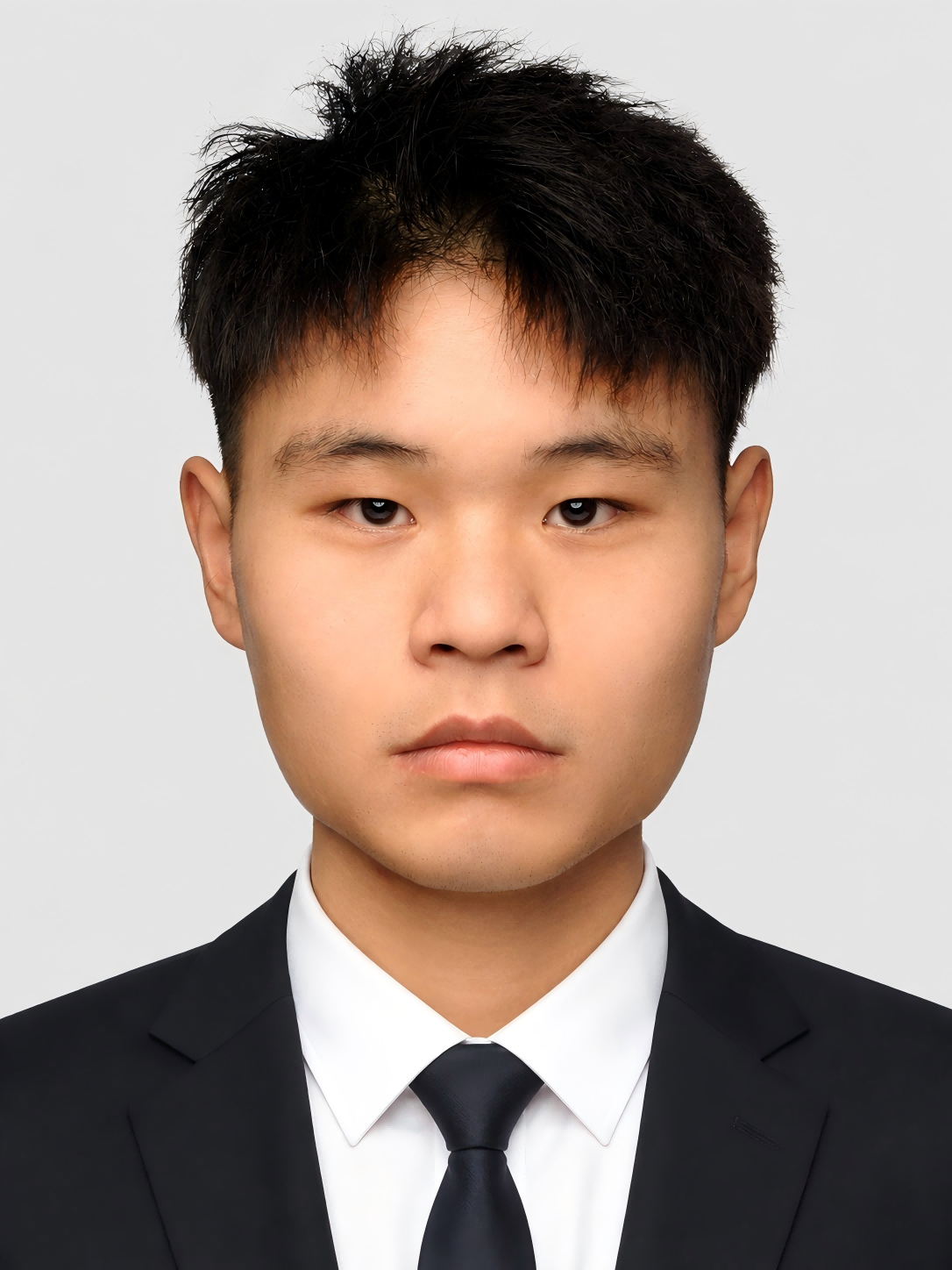}}]{Zuzhi Yang}
 received the B.E. degree in Information Security from Civil Aviation University of China, Tianjin, China, in 2023. He is currently pursuing the M.S. degree with Hangzhou Dianzi University, Hangzhou, China. His research interests include embodied intelligence and 3D computer vision.
\end{IEEEbiography}

\begin{IEEEbiography}[{\includegraphics[width=1in,height=1.25in,clip,keepaspectratio]{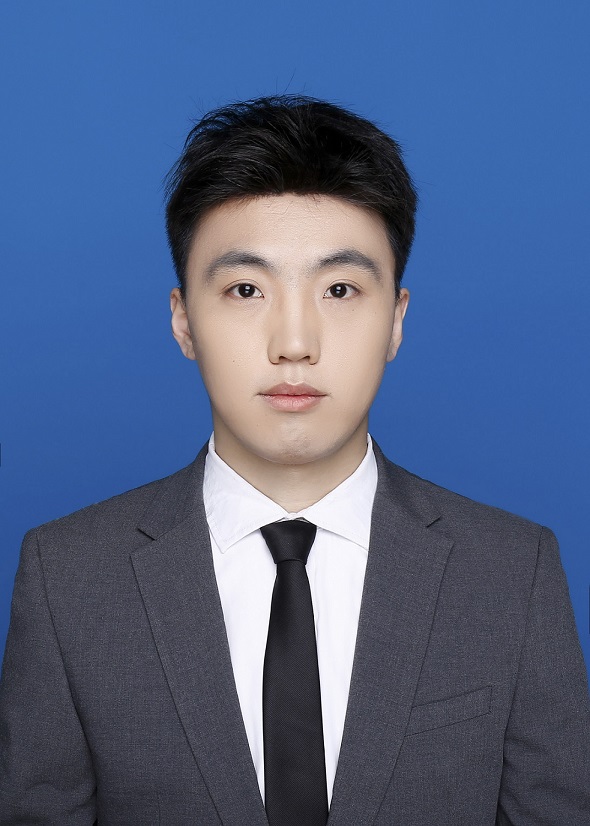}}]{Bingtao Ma}
received the B.S. degree from Hebei University of Technology in 2017, and the Ph.D. degree from the State Key Laboratory of Robotics, Shenyang Institute of Automation, Chinese Academy of Sciences, Shenyang, China, in 2024. He is currently a Specially Appointed Associate Professor and a Master Supervisor with Hangzhou Dianzi University, Hangzhou, China, and a reviewer for IEEE TCSVT. His current research interests include 3D computer vision, transfer learning, and continual learning.
\end{IEEEbiography}

\begin{IEEEbiography}[{\includegraphics[width=1in,height=1.25in,clip,keepaspectratio]{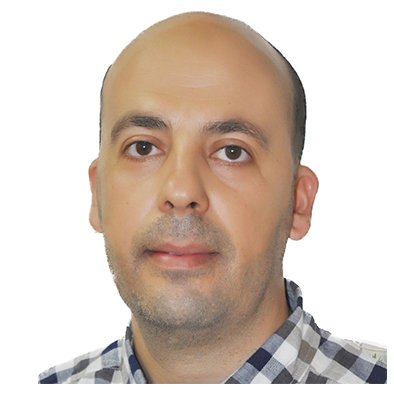}}]{Mounir Kaaniche} (IEEE M'10, SM'20) received the Engineering and Master degrees in telecommunications from SUP'COM-Tunis in 2006 and 2007, and the Ph.D. degree from T\'el\'ecom ParisTech, Paris, France, in 2010. From Jan. 2011 to Sept. 2012, he was a postdoctoral researcher at the Signal and Image Processing Department of T\'el\'ecom ParisTech. From Oct. 2012 to Aug. 2013, he worked as a Graduate Teaching Assistant with Universit\'e Paris-Est Marne-la-Vall\'ee, France. From Sept. 2013 to Aug. 2025, he was an Associate Professor at Universit\'e Sorbonne Paris Nord (USPN) and member of the Multimedia team of Laboratoire de Traitement et Transport de l’Information. Since Sept. 2025, he is Full Professor at USPN. Since Sept. 2022, he is also associate researcher at the Center for Visual Computing (CVN), CentraleSup\'elec, Universit\'e Paris-Saclay, France. His research interests focus on adaptive wavelets as well as statistical and deep learning models for various 2D and 3D image analysis/processing tasks including compression, classification/retrieval, quality enhancement and quality assessment. 
\end{IEEEbiography}

\begin{IEEEbiography}[{\includegraphics[width=1in,height=1.25in,clip,keepaspectratio]{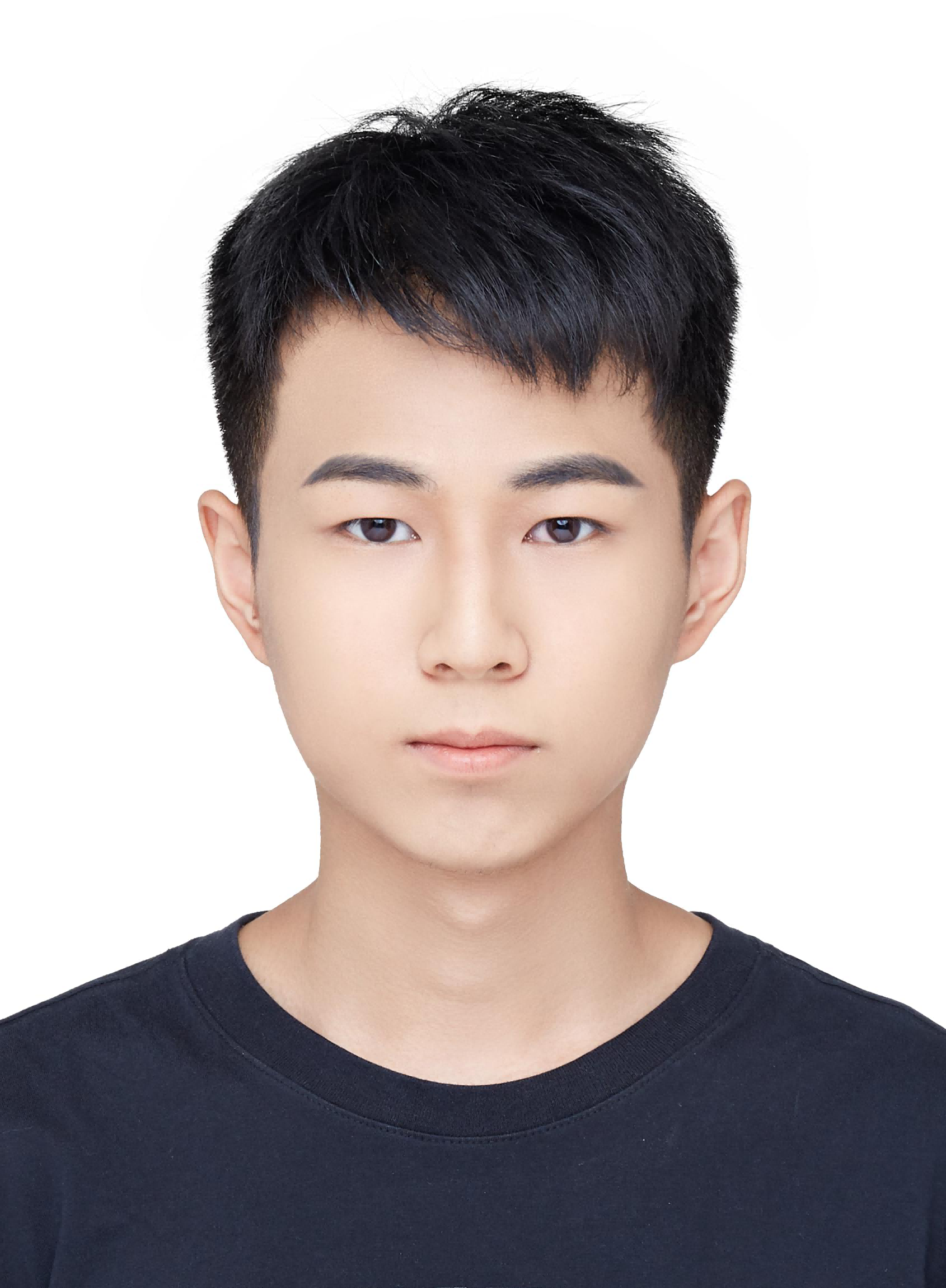}}]{Ziwei Li}
is with the School of Computer Science and Technology, University of Science and Technology of China, Hefei, China.
\end{IEEEbiography}

\begin{IEEEbiography}[{\includegraphics[width=1in,height=1.25in,clip,keepaspectratio]{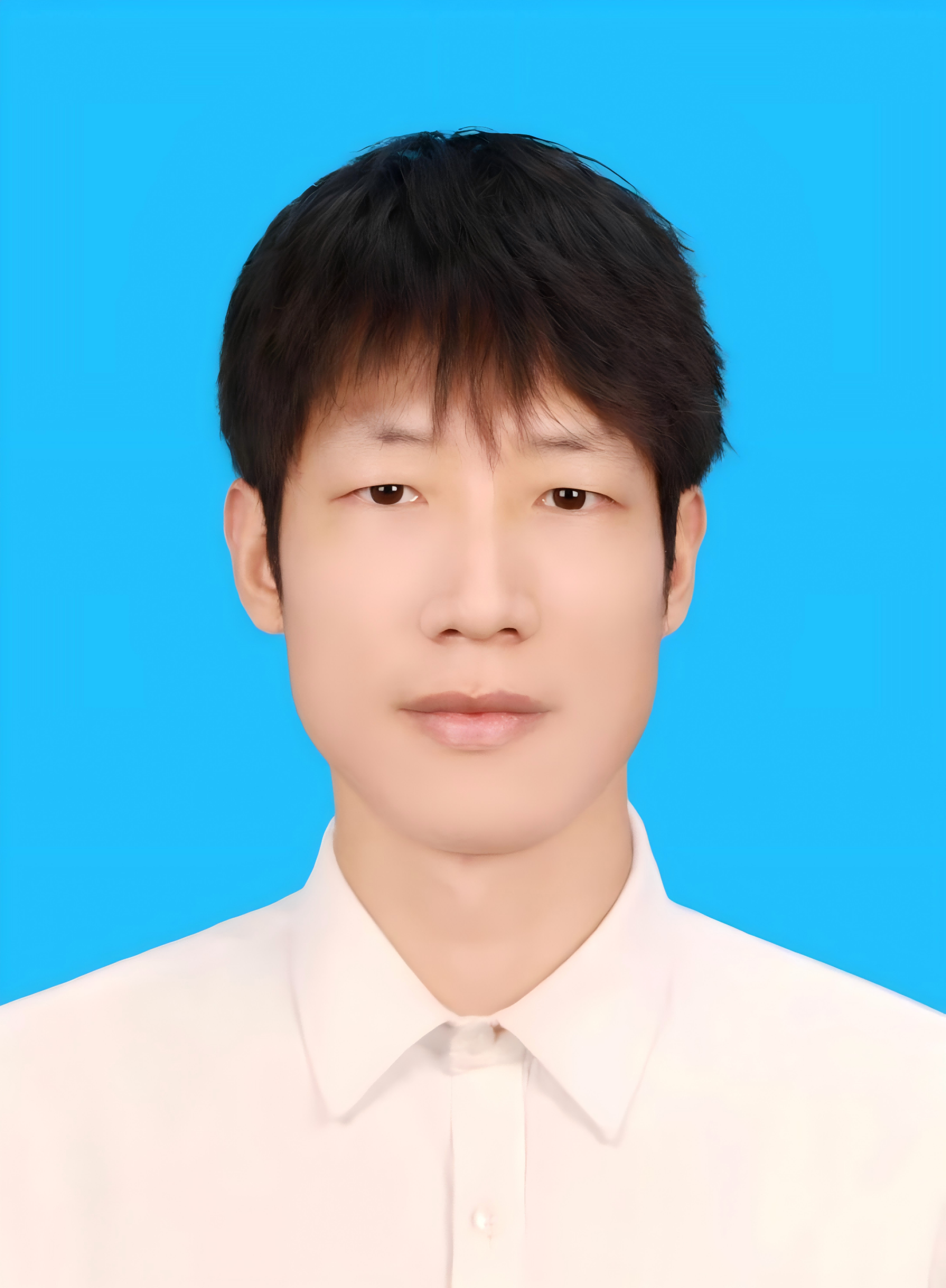}}]{Zhiming Cheng}
received his M.S. degree in software engineering from the University of South China. He received the Ph.D. degree in control science and engineering from Hangzhou Dianzi University in 2024. He is currently an Assistant Researcher with Hangzhou Dianzi University. His research interests include transfer learning, computer vision, and pattern recognition.
\end{IEEEbiography}

\begin{IEEEbiography}[{\includegraphics[width=1in,height=1.25in,clip,keepaspectratio]{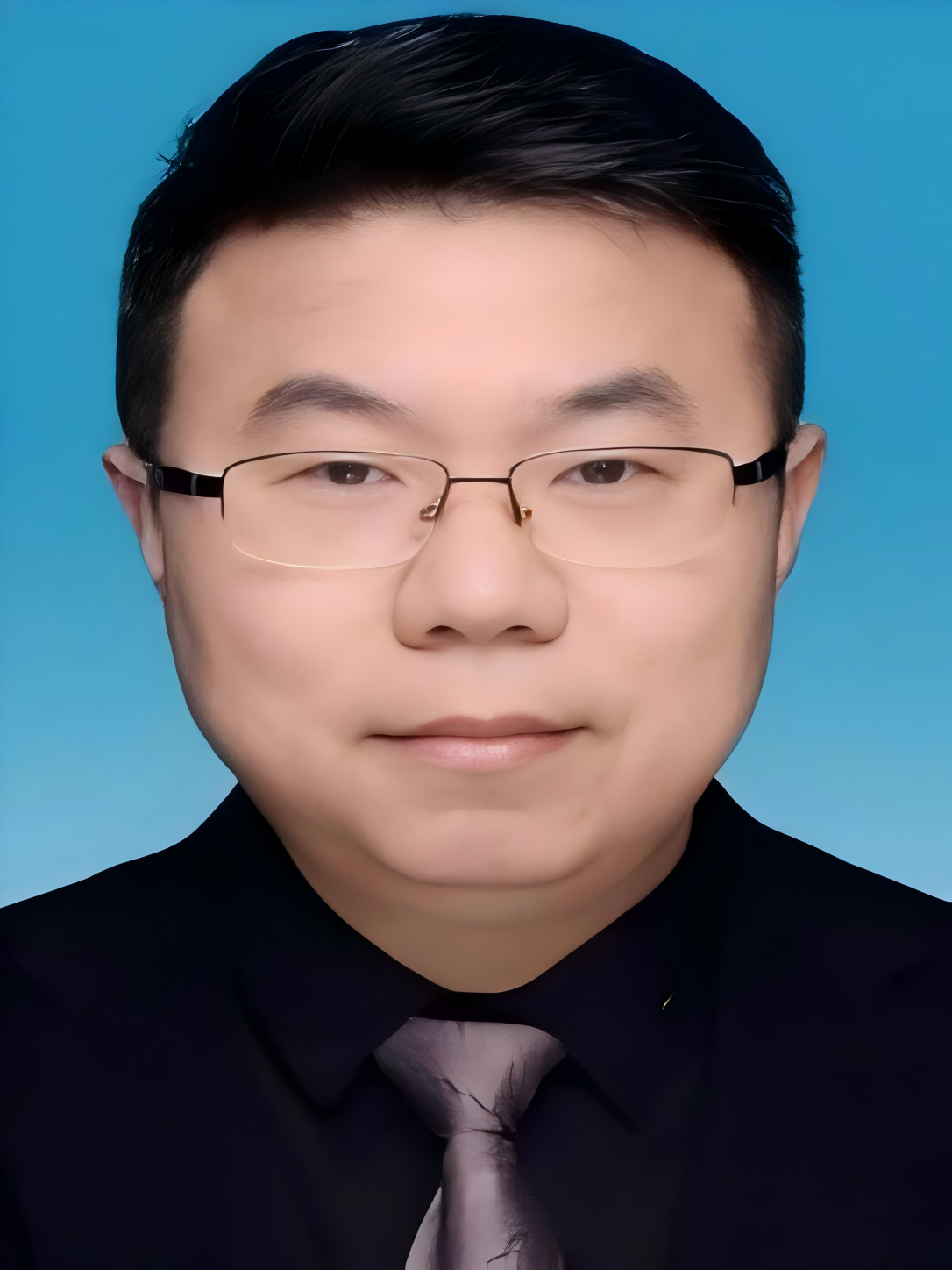}}]{Zhidong Zhao}
received the B.S. and M.S. degrees in mechanical engineering from the Nanjing University of Science and Technology, Nanjing, China, in 1998 and 2001, respectively, and the Ph.D. degree in biomedical engineering from Zhejiang University, Hangzhou, China, in 2004. He is currently a Full Professor with Hangzhou Dianzi University, Hangzhou. His research interests include biomedical signal processing, wireless sensor networks, biometrics, and machine learning.
\end{IEEEbiography}

\begin{IEEEbiography}[{\includegraphics[width=1in,height=1.25in,clip,keepaspectratio]{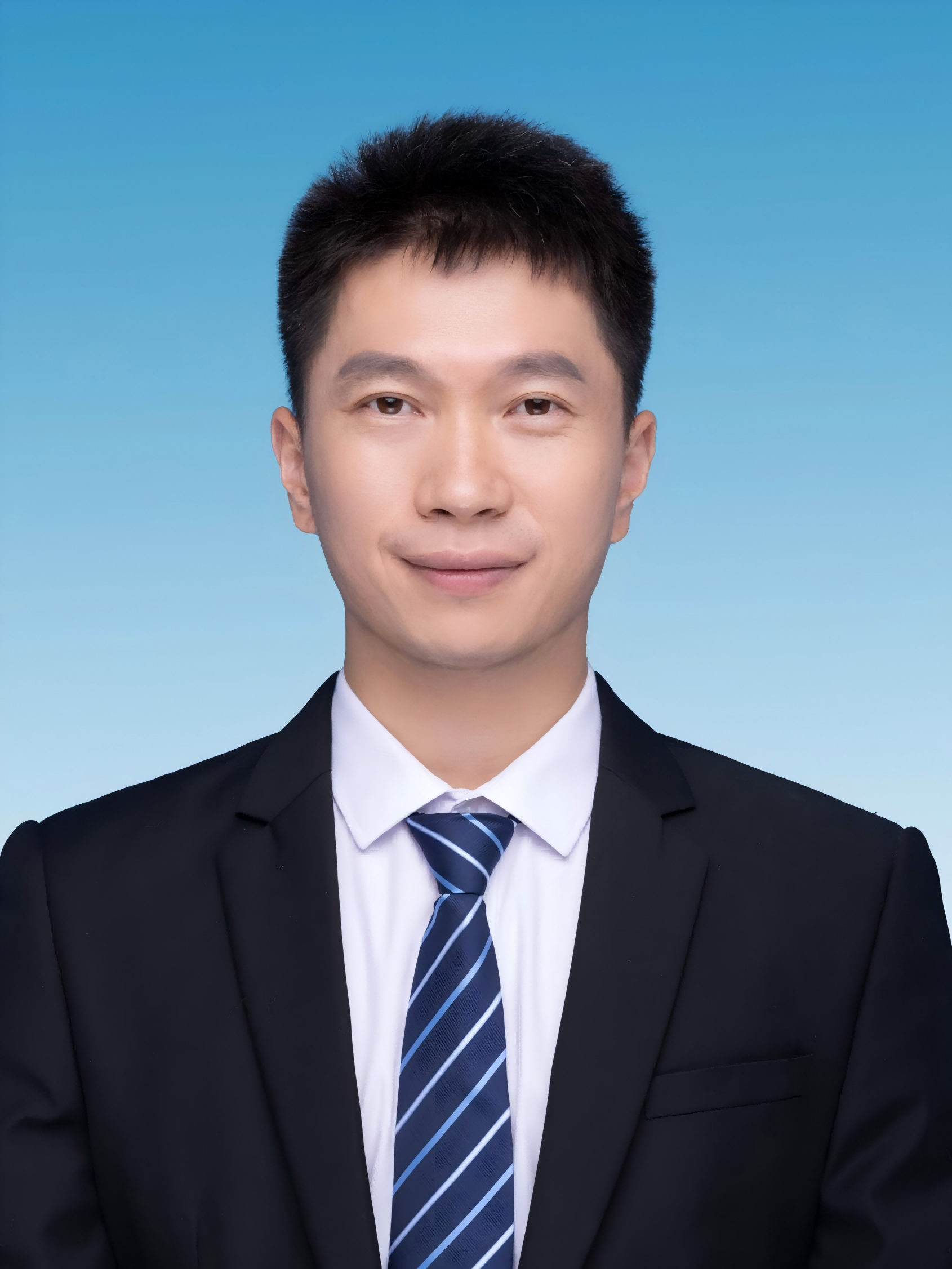}}]{Chenggang Yan}
received the B.S. degree in computer science from Shandong University in 2008 and the Ph.D. degree in computer science from the Institute of Computing Technology, Chinese Academy of Sciences, in 2013. He is currently a Professor with Hangzhou Dianzi University. Before that, he was an Assistant Research Fellow with Tsinghua University. His research interests include intelligent information processing, machine learning, image processing, computational biology, and computational photography. He is currently serving as an Associate Editor of IEEE Transactions on Circuits and Systems for Video Technology (TCSVT).
\end{IEEEbiography}

\begin{IEEEbiography}[{\includegraphics[width=1in,height=1.25in,clip,keepaspectratio]{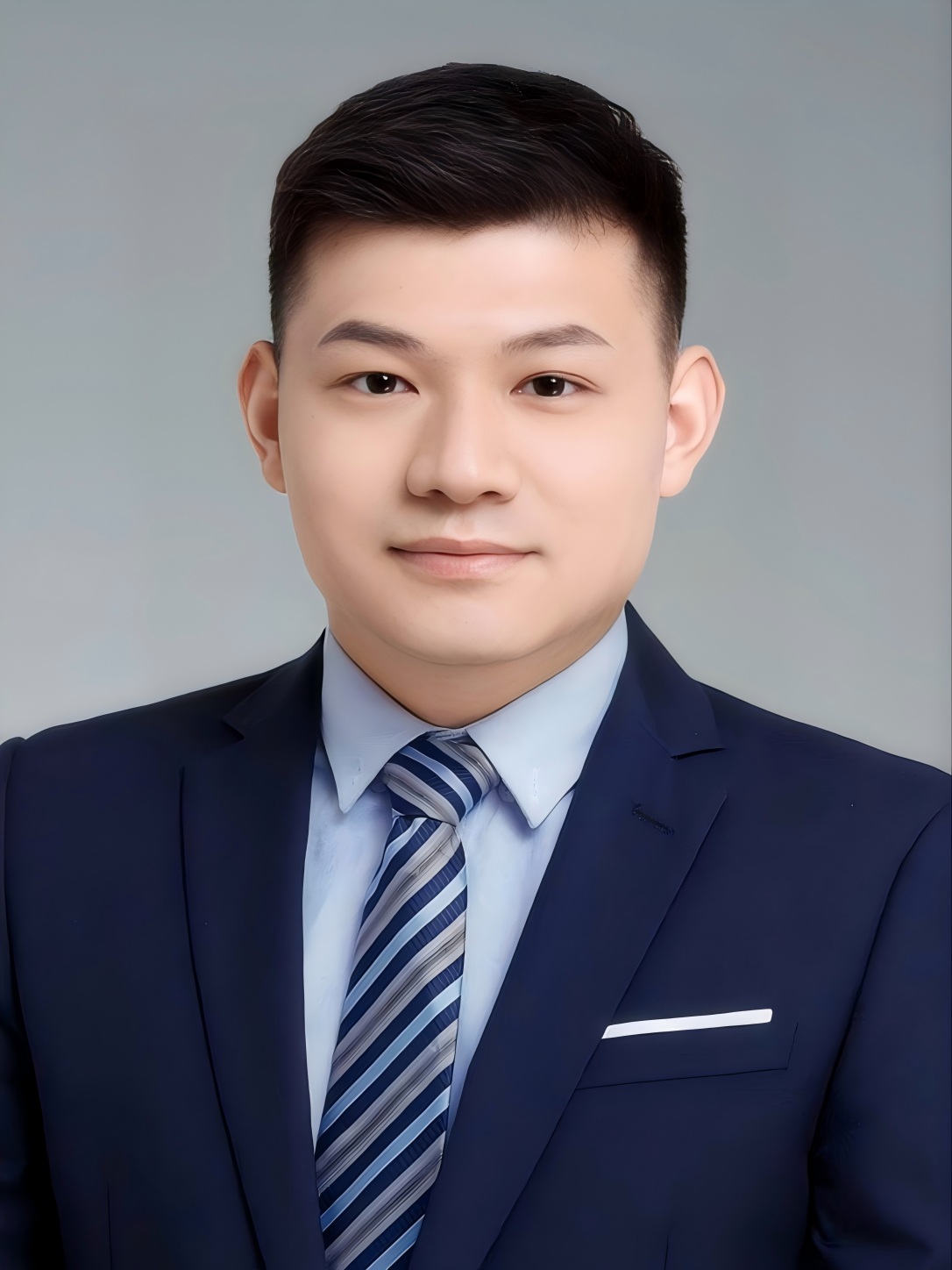}}]{Shuai Wang}
received the BS degree from the China University of Mining and Technology, Xuzhou, in 2010, and the Ph.D. degree from the State Key Laboratory of Robotics, Chinese Academy of Sciences, Shenyang, in 2017. He was a Postdoc Fellow at the School of Medicine, University of North Carolina at Chapel Hill (2017–2019) and a Visiting Fellow at the Clinical Center, National Institutes of Health (2020–2021). His current research interests include medical image analysis, compute vision, machine learning and transfer learning.
\end{IEEEbiography}

\vfill

\end{document}